\documentclass[lettersize,journal]{IEEEtran}

\usepackage{amsmath,amsfonts}

\usepackage{algorithm}
\usepackage{setspace}
\usepackage[noend]{algpseudocode}
\usepackage{tikz}

\usepackage[commentColor=black]{algpseudocodex}
\tikzset{algpxIndentLine/.style={draw=black}}
\algrenewcommand{\alglinenumber}[1]{\bfseries\footnotesize #1}
\algrenewcommand{\algorithmicrequire}{\textbf{Input:}}
\algrenewcommand{\algorithmicensure}{\textbf{Output:}}

\usepackage{array}
\usepackage{multirow, array, ragged2e, lscape, eqparbox, xspace, setspace}
\usepackage{booktabs, tabularx,tabulary}

\usepackage[caption=false,font=normalsize,labelfont=sf,textfont=sf]{subfig}
\usepackage{graphicx}

\usepackage[colorlinks = true,
            linkcolor = blue,
            urlcolor  = blue,
            citecolor = blue,
            anchorcolor = blue]{hyperref}
\usepackage{url}
\usepackage{cite}
\usepackage{orcidlink}

\usepackage{textcomp}
\usepackage{stfloats}
\usepackage{verbatim}
\usepackage{makecell}
\usepackage{pifont}
\usepackage{enumitem}  
\usepackage{textcomp}

\graphicspath{{figures/}}

\usepackage{hyphenat}

\renewcommand{\vec}[1]{\mathbf{#1}}

\let\oldsubsection\subsection
\renewcommand{\subsection}[1]{\vspace{-0.3em}\oldsubsection{#1}}

\makeatletter
\DeclareRobustCommand\onedot{\futurelet\@let@token\@onedot}
\def\@onedot{\ifx\@let@token.\else.\null\fi\xspace}

\def\bb{\mathbb}
\makeatother

\newcommand\copyrighttext{%
  \footnotesize \textcopyright 2024 IEEE. Personal use of this material is permitted.
  Permission from IEEE must be obtained for all other uses, in any current or future
  media, including reprinting/republishing this material for advertising or promotional
  purposes, creating new collective works, for resale or redistribution to servers or
  lists, or reuse of any copyrighted component of this work in other works.
  DOI: \href{https://doi.org/10.1109/TPWRS.2024.3447533}{10.1109/TPWRS.2024.3447533}}
\newcommand\copyrightnotice{%
\begin{tikzpicture}[remember picture,overlay]
\node[anchor=south,yshift=10pt] at (current page.south) {\parbox{\dimexpr\textwidth-\fboxsep-\fboxrule\relax}{\copyrighttext}};
\end{tikzpicture}%
}

\begin{document}

\title{ A Heterogeneous Graph-Based Multi-Task Learning for Fault Event Diagnosis in Smart Grid}

\author{Dibaloke Chanda \orcidlink{0000-0001-5993-659X},~\IEEEmembership{Student Member,~IEEE}, \\ Nasim Yahya Soltani \orcidlink{0000-0002-4502-8715},~\IEEEmembership{Member,~IEEE}
\thanks{Part of this work was presented at the Intl. Workshop on Machine Learning for Signal Processing, Rome, Italy, September 2023.}
\thanks{}}

\markboth{}%
{Shell \MakeLowercase{\textit{ Chanda et al.}}: A Heterogeneous Multi-Task Learning Graph Neural Network for Fault Event Diagnosis in Distribution Smart Grids }

\maketitle

\copyrightnotice

\begin{abstract}

Precise and timely fault diagnosis is a prerequisite for a distribution system to ensure minimum downtime and maintain reliable operation. This necessitates access to a comprehensive procedure that can provide the grid operators with insightful information in the case of a fault event. In this paper, we propose a heterogeneous multi-task learning graph neural network (MTL-GNN) capable of detecting, locating and classifying faults in addition to providing an estimate of the fault resistance and current. Using a graph neural network (GNN) allows for learning the topological representation of the distribution system as well as feature learning through a message-passing scheme. We investigate the robustness of our proposed model using the IEEE-123 test feeder system. This work also proposes a novel GNN-based explainability method to identify key nodes in the distribution system which then facilitates informed sparse measurements. Numerical tests validate the performance of the model across all tasks.    

\end{abstract}


\section{Introduction}\label{sec:Introduction}
\IEEEPARstart{F}{\lowercase{ault}} diagnosis is a crucial task for the operation and maintenance of power systems, particularly in distribution networks due to the nature of complex interconnectivity and scale of the network. Failure to take proper action during a fault event can result in a cascading outage of the distribution system~\cite{veloza2016analysis,guo2017critical}. For uninterrupted operation in the case of a fault occurrence grid operators need to identify the precise location of the fault in addition to the type of the fault. Furthermore, knowing the fault resistance and fault current before taking action for fault isolation and fault clearance guarantees the implementation of appropriate safety measures. This additional information allows grid operators to make more informed decisions and also to plan for necessary repairs or equipment replacements. Not only that, but it also provides insight into post-fault analysis to identify if all the protection systems performed as intended. Due to the advancement in deep learning in recent years, most fault diagnosis systems are utilizing a data-driven approach. However, there is a lack of unified methods that take into account the challenges associated with real-world deployment and many research provides analysis based on theoretical assumptions only.

In this work, we propose a unified heterogeneous MTL-GNN architecture that is capable of performing fault detection, fault localization, fault type classification, fault resistance estimation and fault current estimation. All the tasks are performed in a simultaneous manner as opposed to a sequential manner which ensures the decoupling of tasks. 

We call it a heterogeneous MTL in contrast to a homogeneous MTL due to the fact that the proposed model performs both classification and regression tasks. We take into account all $5$ types of short circuit faults that can occur in a distribution system~\cite{blackburn2006protective}. This includes asymmetrical faults consisting of line-to-ground faults (LG), line-to-line faults (LL), line-to-line-to-ground faults (LLG) and symmetrical faults consisting of line-to-line-to-line-to-ground faults (LLLG), line-to-line-to-line faults (LLL). To address the challenges associated with real-world deployment, our analysis takes into account measurement error, variable fault resistance, small dataset, topology changes and sparse measurements. To make our contribution clear in the following section, the related literature in this domain and the drawbacks and scope for development have been reviewed.

\section{Literature Review} \label{sec:Literature_Review}
The literature on fault event diagnosis is very extensive \cite{bahmanyar2017comparison,dashti2021survey,de2023fault}. This includes different kinds of faults such as over-voltage, insulator, voltage sag, arc, and short-circuit faults. 

They can be broadly structured into two categories. One category are data-driven approaches which utilize deep learning models \cite{javadian2009determining,tokel2018new,guo2017deep,li2019real,zou2022double,souhe2022fault,paul2022knowledge,shadi2022real,chen2019fault,sun2021distribution,de2021fault,li2021ppgn,mo2022sr,hu2022fault,nguyen2023spatial} and another category includes traditional methods \cite{liang2015fault,shi2018travelling,wang2020traveling,tashakkori2019fault,krishnathevar2011generalized,das2012distribution,dashti2014accuracy,jia2016high,gautam2012detection,sekar2017combined,gush2018fault,bayati2021mathematical,wilches2022algorithm,lotfifard2011voltage,dong2013enhancing,trindade2013fault} using statistical measures, signal processing and the physics of fault events. The latter category can be further broken down into separate categories like traveling wave-based\cite{liang2015fault,shi2018travelling,wang2020traveling,tashakkori2019fault}, impedance-based \cite{krishnathevar2011generalized,das2012distribution,dashti2014accuracy,jia2016high}, morphology-based \cite{gautam2012detection,sekar2017combined,gush2018fault,bayati2021mathematical,wilches2022algorithm}, and voltage sag \cite{lotfifard2011voltage,dong2013enhancing,trindade2013fault} methods.

As regards the data-driven approaches, there is a multitude of methods and architectures but we broadly divide them into two categories which are GNN-based approaches 
\cite{chen2019fault,sun2021distribution,de2021fault,li2021ppgn,mo2022sr,hu2022fault,nguyen2023spatial}, multi-layer perceptron~(MLP) and convolutional neural network~(CNN) based approaches\cite{javadian2009determining,tokel2018new,guo2017deep,li2019real,zou2022double,souhe2022fault,paul2022knowledge,shadi2022real}.

\subsection{Traditional Methods}
The traveling wave-based methods~\cite{liang2015fault,shi2018travelling,wang2020traveling,tashakkori2019fault} analyze the characteristics of traveling waves generated when a fault event occurs. During a fault event, the sudden change in voltage and current around the fault location initiates a transient disturbance that propagates along the distribution line.  This transient wave and its reflected counterpart are picked up by fault recorders installed at a substation or two substations depending on whether it is a single-end method or a double-ended method. Based on the time of arrival of the transient wave and the reflected wave it is possible to deduce the location of the fault.

\begin{table*}[t]
\centering
\caption{Summary of Technical Differences with Previous GNN-Based Literature}
    \label{differences_with_literatures}
    \renewcommand{\arraystretch}{1}
    \resizebox{1.6\columnwidth}{!}{
    \setlength{\extrarowheight}{1em}
\begin{tabular}{cllllllllllllllll}
\toprule
\bf Tasks &  & \bf \makecell[c]{Chen et al.~\cite{chen2019fault}} &  &  \makecell[c]{\bf Sun et al.~\cite{sun2021distribution}} &  &  \makecell[c]{ \bf Freitas et al.~\cite{de2021fault}} &  & \makecell[c]{\bf Li et al.~\cite{li2021ppgn}} &  &  \makecell[c]{\bf Mo et al.~\cite{mo2022sr}} &  & \makecell[c] {\bf Hu et al.~\cite{hu2022fault}} &  & \makecell[c]{\bf Nguyen et al.~\cite{nguyen2023spatial}} &  & \makecell[c]{\bf Ours} \\ \cline{1-1} \cline{3-3} \cline{5-5} \cline{7-7} \cline{9-9} \cline{11-11} \cline{13-13} \cline{15-15} \cline{17-17}

\multicolumn{1}{c}{\bf  \makecell[c]{Fault Resistance}}& &  \makecell[c]{ \bf Variable} &  &  \makecell[c]{ \bf Variable} &  &  \makecell[c]{ \bf Variable} &  & \makecell[c]{ \bf Variable}  &  &  \makecell[c]{ \bf Constant} &  &  \makecell[c]{ \bf Constant} &  & \makecell[c]{ \bf Constant}  &  & \makecell[c]{ \bf Variable}   \\ 

\bottomrule

\begin{tabular}[c]{@{}c@{}} \makecell[c]{Fault \\ Detection}\end{tabular}            &  & \makecell[c]{\ding{55}}  &  & \makecell[c]{\ding{55}} &  &  \makecell[c]{\ding{55}}  &  & \makecell[c]{\ding{55}} &  & \makecell[c]{\ding{55}}  &  &  \makecell[c]{\ding{55}} &  & \makecell[c]{\ding{51}} &  & \makecell[c]{\ding{51}} \\ \hline

\begin{tabular}[c]{@{}c@{}} \makecell[c]{Fault \\ Localization}\end{tabular}          &  &  \makecell[c]{\ding{51}} &  & \makecell[c]{\ding{51}} &  &  \makecell[c]{\ding{51}} &  & \makecell[c]{\ding{51}} &  & \makecell[c]{\ding{51}}  &  & \makecell[c]{\ding{51}}  &  & \makecell[c]{\ding{51}}  &  &  \makecell[c]{\ding{51}} \\ \hline

\begin{tabular}[c]{@{}c@{}}\makecell[c]{Fault \\  Classification}\end{tabular}       &  &  \makecell[c]{\ding{55}} &  & \makecell[c]{\ding{55}} &  & \makecell[c]{\ding{55}}  &  & \makecell[c]{\ding{55}}  &  & \makecell[c]{\ding{51}}  &  & \makecell[c]{\ding{51}} &  & \makecell[c]{\ding{51}} &  &  \makecell[c]{\ding{51}} \\ \hline

\begin{tabular}[c]{@{}c@{}} \makecell[c]{Fault Resistance \\ Estimation}\end{tabular} &  &  \makecell[c]{\ding{55}} &  &  \makecell[c]{\ding{55}} &  &  \makecell[c]{\ding{55}} &  & \makecell[c]{\ding{55}} &  & \makecell[c]{\ding{55}}  &  & \makecell[c]{\ding{55}}  &  & \makecell[c]{\ding{55}} &  & \makecell[c]{\ding{51}} \\ \hline

\begin{tabular}[c]{@{}c@{}} \makecell[c]{Fault Current \\ Estimation}\end{tabular}   &  &  \makecell[c]{\ding{55}}  &  &  \makecell[c]{\ding{55}} &  & \makecell[c]{\ding{55}}  &  & \makecell[c]{\ding{55}} &  & \makecell[c]{\ding{55}} &  & \makecell[c]{\ding{55}} &  & \makecell[c]{\ding{55}} &  & \makecell[c]{\ding{51}} \\

\bottomrule

\begin{tabular}[c]{@{}c@{}} \makecell[c]{ \it Types of \\ \it Fault Considered}\end{tabular}   &  &  \makecell[c]{ \it LG, LL, LLG}  &  &\makecell[c]{ \it LG, LL, LLG \\  \it LLL, LLLG}   &  & \makecell[c]{ \it LG} &  & \makecell[c]{ \it LG, LL, LLG} &  & \makecell[c]{ \it LG, LL, LLG}  &  &  \makecell[c]{ \it LG, LL, LLG \\ \it LLL, LLLG }&  &  \makecell[c]{ \it LG, LL, LLG \\ \it LLL, LLLG } &  & \makecell[c]{ \it LG, LL, LLG \\ \it LLL, LLLG } \\

\bottomrule

\end{tabular}
}
\end{table*}

In impedance-based methods~\cite{krishnathevar2011generalized,das2012distribution,dashti2014accuracy,jia2016high}, current and voltage signals are measured along different places on the distribution line. In the case of a fault event, these measured signals are used to isolate the fundamental frequency to estimate the apparent impedance. This apparent impedance is then used to locate the fault event.

 Morphology-based methods~\cite{gautam2012detection,sekar2017combined,gush2018fault,bayati2021mathematical,wilches2022algorithm}  make use of mathematical morphological operations like dilation, erosion, closing and opening on the waveform generated during a fault event to extract features that correspond to a fault event. After feature extraction is complete, the extracted features are passed to different classifier algorithms like decision trees~\cite{sekar2017combined}, recursive least square stage~\cite{gush2018fault}, and random forest~\cite{wilches2022algorithm}.
 
Voltage sag methods~\cite{lotfifard2011voltage,dong2013enhancing,trindade2013fault} use the characteristic of the reduction of voltage magnitude in the case of a fault event to isolate the fault's exact location. When a fault event occurs there is a sudden dip in the voltage magnitude. This sudden dip occurs only at the location of the fault event which can isolated based on the characteristics of the voltage sag.

\subsection{Deep Learning Based Methods}
\subsubsection{MLP and CNN Based Methods} These methods use historical or software-simulated data relating to fault events in distribution systems and use them to train MLP and CNN architectures to do prediction tasks like fault detection, fault localization and fault classification. 

The work in \cite{javadian2009determining} is one of the early papers that use a MLP. The input to their proposed model is the current measures of distributed generation units~(DGs) and substation and they perform fault localization as output. Another similar work that uses MLP is \cite{tokel2018new} where the authors use the IEEE-13 bus system to perform both fault classification and localization. In~\cite{souhe2022fault} the authors use MLP but with an additional fuzzy layer and their analysis is on the IEEE-37 bus system.

In~\cite{guo2017deep} the CNN architecture is adopted for fault localization. First, the authors use a continuous wavelet transform (CWT) algorithm to convert current phasors to images which are fed to a CNN model to localize the fault. The work in~\cite{li2019real} also uses a CNN to localize faults but considers partial observability of the grid on IEEE-39 and IEEE-68 bus systems. In~\cite{zou2022double} the authors take a slightly different approach by using 1-D convolutions with double-stage architecture. The first stage extracts the features and the second performs fault identification. Similar to~\cite{guo2017deep}, the work in~\cite{paul2022knowledge} also uses CWT to convert time-domain current signals to image domain and use the transformed data for fault classification and localization on the IEEE-34 bus system. A different approach by using a capsule-based CNN is proposed in~\cite{shadi2022real} to do fault detection, localization and classification.

\subsubsection{GNN Based Methods}

The first prominent work to use GNN for fault localization is in~\cite{chen2019fault}. In this work, IEEE-123 and IEEE-37 are used as the test feeder system and the authors consider a range of factors like metering error, changes in topology, etc. for their analysis. The architecture employed is CayleyNets~\cite{levie2017cayleynets} which is a graph convolutional neural network (GCN) based on spectral theory. The feature vector considered as the input to their model consists of both voltage and current phasors measured from the buses in the distribution system.

The subsequent notable work is~\cite{sun2021distribution} which utilizes not only node features but also link features that include branch impedance, admittance and
regulation parameters of the distribution lines. The authors validate their approach on a self-designed $6.6$~KV system with $12$~buses and $8$~loads.

Two other related research works are in~\cite{de2021fault} and~\cite{li2021ppgn}. For the first one, the authors consider a gated GNN~\cite{li2015gated} architecture. However, they only consider single-line-to-ground fault as opposed to~\cite{li2021ppgn} which considers all three types of asymmetrical short circuit fault. Not only that, but the authors also consider limited observability and limited labels in their implementation. Their proposed method has a two-stage architecture with only voltage phasors as input.

In~\cite{mo2022sr} there is a more recent work that uses a different variant of GNN, a graph attention neural network~(GAT)~\cite{velivckovic2017graph} to do both fault localization and classification in IEEE-37 feeder system. In their analysis, the authors consider a constant fault resistance and the fault localization prediction is dependent on the fault classification task.

These above-mentioned works consider instantaneous current and/or voltage phasors meaning that they require measurement at the fault time, no pre-fault or post-fault measurement is required. But the method proposed in~\cite{hu2022fault} requires a fault waveform sampled at $1$ KHz. Similar to~\cite{mo2022sr} their analysis considers constant fault resistance. One important contribution of this research work is that they use MTL to do both fault classification and fault localization at once in contrast to \cite{mo2022sr}.

Another similar recent work employs spatial-temporal recurrent GNN to do three tasks simultaneously which are fault detection, classification and localization ~\cite{nguyen2023spatial} . The authors report numerical results tested on a microgrid and IEEE-123 bus system. Similar to the previous approach their analysis considers constant resistance and due to the temporal nature of their proposed method, it requires high time resolution of fault waveshape. 

The summary of major technical differences between our proposed model with the existing GNN-based models is outlined in Table~\ref{differences_with_literatures}. We make the argument that the models that are only trained for fault localization and/or classification \cite{chen2019fault,sun2021distribution,de2021fault,li2021ppgn,mo2022sr,hu2022fault} will require a separate method to first distinguish between fault event and other events (non-fault) in distribution system like load change.

Also, the models that consider constant resistance\cite{mo2022sr,hu2022fault,nguyen2023spatial} will only perform well as long as the actual fault resistance is similar to the fault resistance considered during the generation of the training dataset. Even though these research work report their model performance with different resistance values, these reported values are only applicable to that specific resistance value. In contrast, \cite{chen2019fault,sun2021distribution,de2021fault,li2021ppgn} takes a more practical approach and train their model considering a range of possible fault resistance values. As long as the fault resistance is within that range, the model is expected to hold its performance.

The drawback of \cite{hu2022fault,nguyen2023spatial} is that their proposed model is dependent on temporal characteristics.  To perform well the dataset needs to have high temporal resolution. As the scale of the distribution system grows data acquisition, storage and training overhead for this approach becomes progressively more demanding. In addition, there is a requirement for perfect time synchronization which further complicates the system design. 

Considering these drawbacks the key contributions of our work are outlined as:

\begin{enumerate}
    \item We propose a unified heterogeneous MTL-GNN architecture to perform $5$ different tasks simultaneously for a fault event which are fault detection, fault localization, fault type classification, fault resistance estimation and fault current estimation.

    \item The proposed model performs well in the presence of measurement error, variable resistance and topology changes as common factors considered in real-world deployments.

    \item We utilize an explainability algorithm specific to GNN to identify key nodes in the grid which provides the opportunity for informed sparse measurement.

\end{enumerate}

The remaining parts of the paper
are organized as follows. In section~\ref{sec:Dataset_Generation}, a brief overview of the test feeder system and the process involving the dataset generation is provided. The mathematical framework for the overall methodology and architecture used is given in section~\ref{sec:Mathematical_Framework}. In the following section~\ref{sec:Architecture_Detail_and_Model_Training}, we outline the details of the architecture, training procedure and hyperparameters assumed during training. Numerical results and discussions on them are presented in section~\ref{sec:Numerical_Tests}. Finally, section~\ref{sec:Conclusion_and_Scope_for_Future_Work} concludes the paper. 

\section{Dataset Generation} \label{sec:Dataset_Generation}

This section briefly covers the details of the IEEE-123 node feeder system and the dataset generation process as well as the underlying assumptions considered in the generation process.

\subsection{IEEE-123 Node Feeder System}
\vspace{-1em}

\begin{figure}[thbp]
    \centering
    \includegraphics[width=0.9\linewidth]{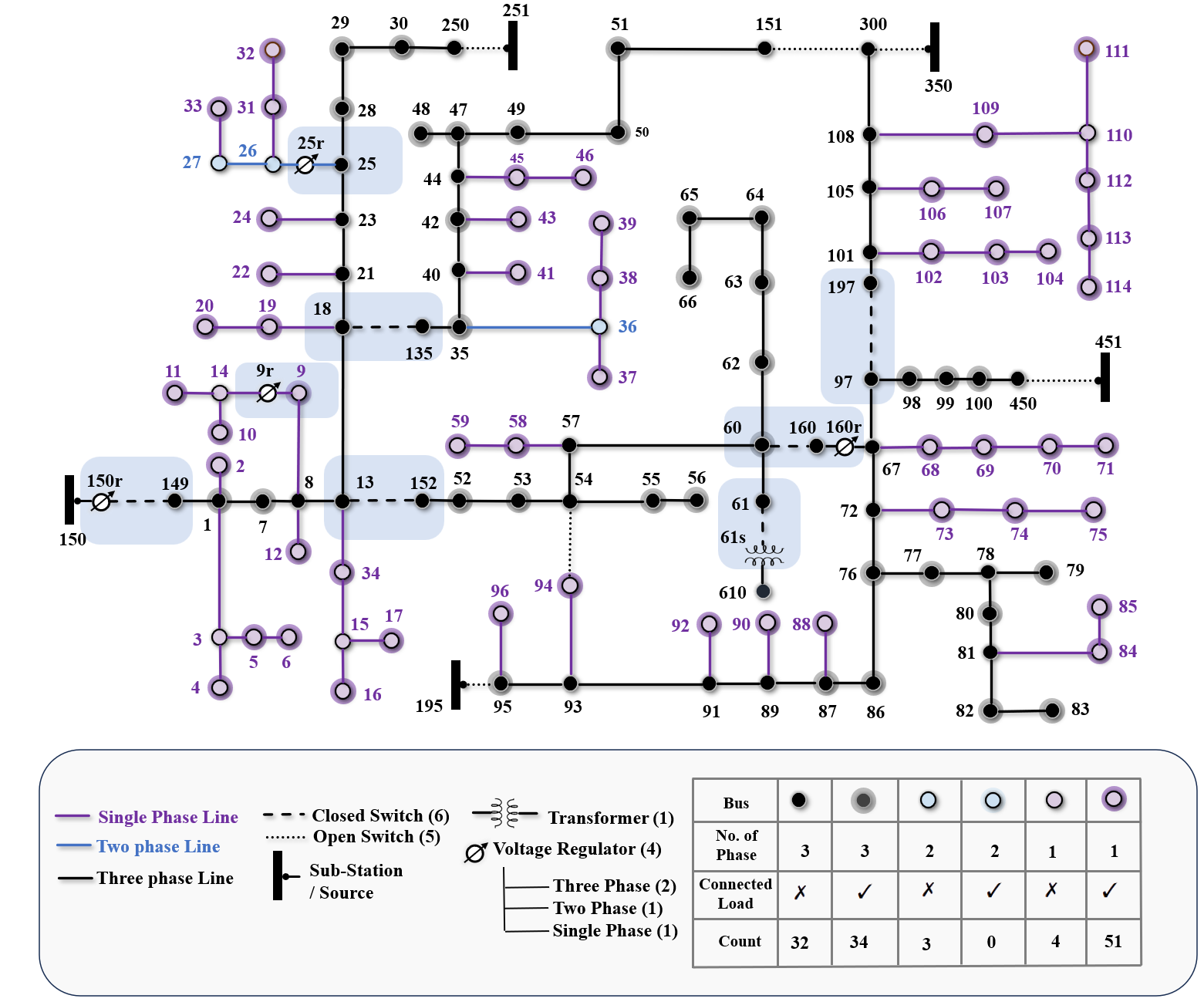}
    \caption{Diagram of IEEE-123 node feeder system. The highlighted blue blocks represent the node pairs that are considered connected. The number of voltage regulators, transformers, and switches is mentioned in ($\cdot$) and the number of buses, their phases and load connectivity are mentioned in the table. The active substation (source bus) is connected to the node $150r$.}
    \label{fig:ieee_123_feeder_system}
\end{figure}

The IEEE-123 node feeder system~\cite{kersting1991radial} shown in Fig.~\ref{fig:ieee_123_feeder_system} operates at a nominal
voltage of $4.16$ KV and consists of both overhead and underground lines. It has three-phase, two-phase and single-phase lines and a couple of open and closed switches, voltage regulators and a transformer.

There are a total of $85$ nodes that have loads connected to them and most of them are connected to single-phase buses and the rest are connected to three-phase buses. For simulating all $5$ types of short circuit faults including asymmetrical and symmetrical faults all three phases need to be considered. Therefore, in our analysis, similar to \cite{chen2019fault,li2021ppgn} we only consider three-phase nodes which tally up to $68$ nodes including the $2$ three-phase regulators $150$r and $160$r. Also, similar to \cite{chen2019fault,li2021ppgn} we also make the assumption that some specific pairs of nodes are connected which are ($149$, $150$r), ($18$, $135$), ($13$, $152$), ($60$,  $160$, $160$r,), ($61$, $61$s), ($97$, $197$), ($9$, $9$r), ($25$, $25$r). The reason behind this assumption is the pairs  ($18$, $135$), ($13$, $152$), ($61$, $61$s), ($97$, $197$), ($60$,  $160$) are connected via closed switches and the pairs ($149$, $150$r), ($9$, $9$r), ($25$, $25$r) consists of buses and their corresponding regulators. This is shown in Fig.~\ref{fig:ieee_123_feeder_system} via highlighted blue sections. One important thing to note here, in actuality, considering all the $4$ regulators as separate nodes the total number of nodes in the feeder system results in $128$ nodes.
   
\subsection{Dataset Description and Generation Procedure}\label{dataset_2}

For dataset generation, we use OpenDSS\cite{dugan2011open}, open-sourced by the electric power research institute (ERPI) as a power flow equation solver engine and use \textit{py\_dss\_interface} module in Python to interface with it. 

We opt to utilize only voltage phasors as measurements based on \cite{chen2019fault}, where the authors showed that the performance of their model is almost identical with or without current phasors. Therefore, there is no incentive to use current phasors as features because that would just double the amount of computation needed at the expense of no performance increase. For the three-phase buses shown in Fig.~\ref{fig:ieee_123_feeder_system} for all three phases ( Phase A, Phase B, Phase C) voltage amplitude and angle (in radian) can be measured. For single-phase and two-phase buses values for the missing buses are padded with zero.

\begin{figure}[!h]
    \centering
    \includegraphics[width=0.6\linewidth]{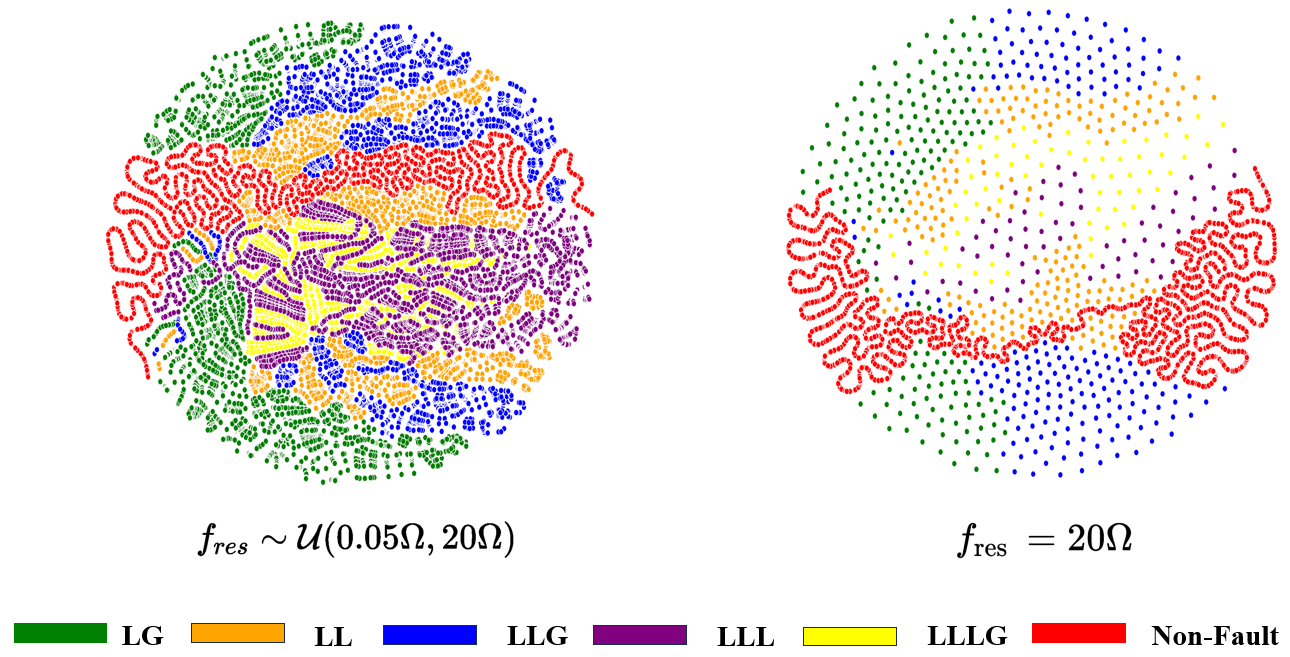}
    \caption{ t-Distributed Stochastic Neighbor Embedding (t-SNE) visualization of all the data points. (\textbf{Left}) shows the dataset generated with a variable range of fault resistance sampled from a uniform distribution $\mathcal{U}(0.05 \Omega, 20 \Omega$). (\textbf{Right}) shows the dataset generated with a constant fault resistance $20 \Omega$.}
    \label{fig:tsne_dataset}
\end{figure}

\begin{figure}[htbp]
    \centering
    \includegraphics[width=0.9\linewidth]{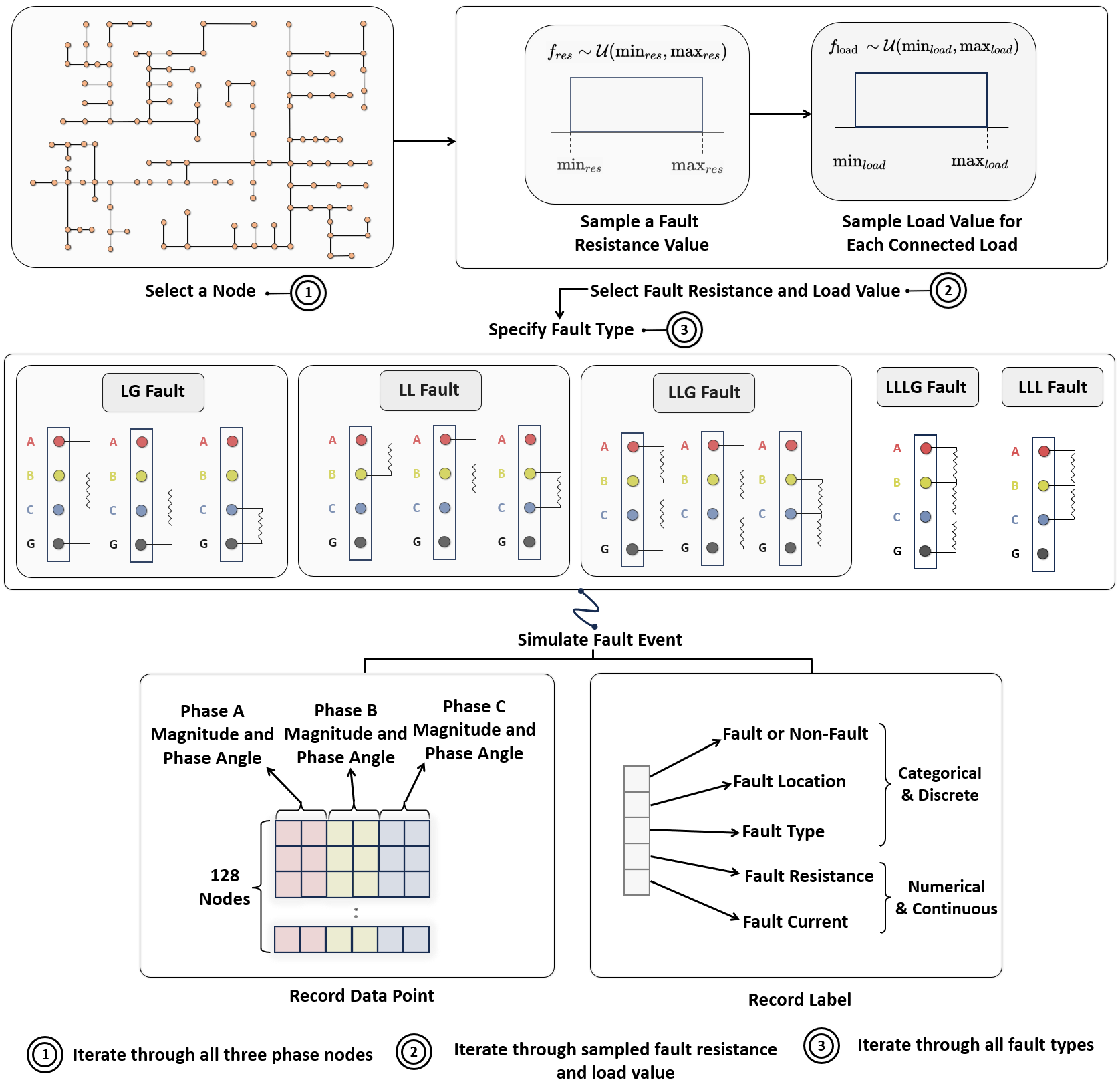}
\caption{Visualization of the sequence of procedures for a single data point and label generation. The double-circled digits represent the loop iteration points in the algorithm. By performing all the iterations in a hierarchical manner (the digits specify the order of the hierarchy) the entire dataset is generated.}
\label{fig:dataset_gen}
\end{figure}

As our proposed model does $5$ tasks simultaneously, for each data point we generate $5$ labels as fault detection, fault location, fault classification, fault resistance, and fault current labels.
The first three labels are discrete values whereas the last two labels are continuous values. As mentioned before for practical consideration we assume a range of fault resistance instead of a single fault resistance value. Fault resistance values are sampled from a uniform distribution (rounded up to $2$ decimal point) which is $f_{res} \sim \mathcal{U}(\min_{res},\max_{res})$, where $f_{res}$ is the sampled fault resistance value and $\min_{res}$ and $\max_{res}$ are the lower and upper bound of the uniform distribution. For our initial analysis, we assume a lower bound of $0.05~\Omega$ and an upper bound of $20~ \Omega$. The practical consideration claim we make is justified by Fig.~\ref{fig:tsne_dataset} which clearly shows that for constant resistance analysis\cite{mo2022sr,hu2022fault,nguyen2023spatial} the fault diagnosis becomes trivial. Fig.~\ref{fig:tsne_dataset} shows t-SNE visualization of the high-dimensional features by projecting them into a 2-dimensional domain. For constant resistance of $20 \Omega$, the features have minimal overlap and there is considerable distance between two neighboring points. Hence, the underlying data distribution is easily separable with non-linear decision boundaries which can be easily learned using any ML method. In our approach with variable fault resistance, the overlap between features is much more pronounced which makes it a much harder prediction task. However, in case of real-world deployment, our proposed approach with the variable fault resistance is much more practical.    

For each fault type, we generate $20400$ data samples. This results in a total of $20400\times5=102000$ data points for the fault events where there are $300$ samples per bus. For non-fault event data generation, we vary all $91$ loads connected to the $85$ buses according to another uniform distribution given by $f_{load} \sim \mathcal{U}(\min_{load},\max_{load}) $. We assume $\min_{load}=20~\text{KW}$ and $\max_{load}=80~\text{KW}$ based on the typical load range associated with the IEEE-123 node feeder system. For non-fault events, the number of data points is also $20400$. The sequence of steps for the data generation process is visualized in Fig.~\ref{fig:dataset_gen}. The double-circled digits in the diagram specify the iteration hierarchy.

\begin{enumerate}
    \item At the first iteration point, we iterate through all $68$ three phase nodes and for a specific node execute the following two steps.
    
    \item At the second iteration point, for that specific node a fault resistance value $f_{res}$ and load values $f_{load}$ for the connected loads are sampled. In actual implementation, this is just iterating through a list consisting of $300$ tuples of $f_{res}$ and $f_{load}$ values that were sampled in advance.
    
    \item  At the third iteration point, we iterate through all $5$ fault types and fault is simulated for each fault type.
\end{enumerate}

Going through the above-mentioned process results in $68\times300\times5=102000$ data points for fault events with associated labels. It should be pointed out that the fault simulation strategy for the asymmetric fault types and symmetric fault types differs slightly. For symmetric fault types which are LG, LL and LLG, the samples consist of $100$ samples for $3$ different states (based on connection difference between phases) which results in $3\times100=300$ samples for each of them.

All the values in the feature vector are standardized by subtracting the mean value and dividing by the standard deviation.
\section{Mathematical 
Framework}\label{sec:Mathematical_Framework}

\begin{figure*}[htbp]
    \centering
    \includegraphics[width=0.8\linewidth]{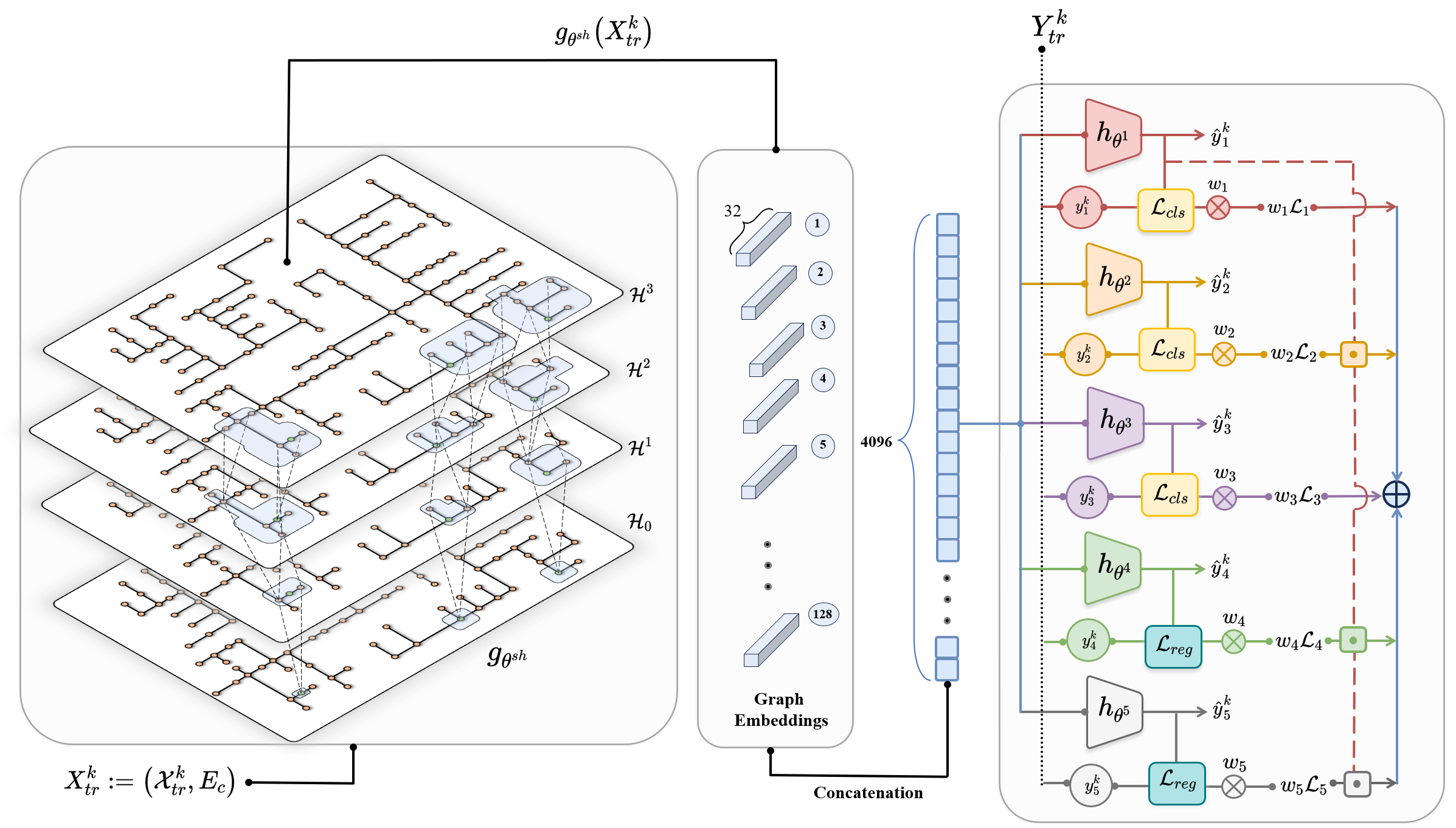}
    \caption{Architecture of the proposed heterogenous MTL-GNN. The input features go through the common backbone GNN to generate graph embeddings. For visual clarity message passing across the layers for a couple of nodes (highlighted in green) is shown, where the blue highlighted sections signify nodes included in the message passing process (\textbf{Left}). The embeddings generated by $128$ nodes are flattened and concatenated together to convert to a one-dimensional vector (\textbf{Middle}). The concatenated feature vector is passed to $5$ heads, three classification heads and two regression heads. The corresponding loss is computed based on the predicted output ($\hat{y}_{t}^{k}$) and ground truth label ($y_{t}^{k}$). The computed loss for each task is weighted and summed together (\textbf{Right}).}
    \label{fig:arch}
\end{figure*}

GNN uses message passing between nodes in the graph for the propagation of features which allows feature representation learning in addition to topological representation learning. GCN is a specific variant of GNN, first introduced by \cite{kipf2016semi}.
The proposed model consists of GCN layers as a common backbone followed by dense layers as prediction heads. GCN perform message passing between a given target node $u$ in a particular layer $l$ and the neighbors of the target node $v \in \mathcal{N}(u)$ according to the following equation:-

\begin{equation}\label{eqn:gcn1}
\mathbf{h}_u^{(l)}=\sigma\left(W^{(l)} \sum_{v \in \mathcal{N}(u) \cup\{u\}} \frac{\mathbf{h}_v}{\sqrt{\lvert \mathcal{N}(u)||\mathcal{N}(v)\rvert}}\right)
\end{equation}

where, $\frac{1}{\sqrt{\lvert \mathcal{N}(u) \mid \mid \mathcal{N}(v) \rvert }}$ is the normalization factor, $\mathbf{h}_{v}$ is the hidden representation of the neighboring nodes, $W^{(l)}$ is the weight matrix consisting trainable parameters, $\sigma$ is the non-linear activation function and  $\mathbf{h}^{(l)}_{u}$ is the hidden representation of the target node. This equation gives an intuitive understanding of how the message passing algorithm works, but for implementation \eqref{eqn:gcn2} is more common in literature.
\begin{equation}\label{eqn:gcn2}
\mathcal{H}^{(l+1)}=\sigma\left(\tilde{D}^{-\frac{1}{2}} \tilde{ \mathcal{A}} \tilde{D}^{-\frac{1}{2}} \mathcal{H}^{(l)} W^{(l)}\right)
\end{equation}
where, $\tilde{\mathcal{A}}=\mathcal{A}+I$ is the adjacency matrix with self-loops and $\mathcal{A}$ is the original adjacency matrix without self-loops,  $\tilde{D}$ is the degree matrix considering the modified adjacency matrix  $\tilde{\mathcal{A}}$ and $\mathcal{H}^{(l)}$ is the $l^{th}$ GCN layer containing all the feature representation of all the nodes for that particular layer. 

The node feeder system is represented by a graph $\mathcal{G}:=(\mathcal{V}, \mathcal{E}, \mathcal{A})$ where $\mathcal{V}$ represents the set of nodes in the feeder system which can be represented as the union of three disjoint sets as shown in the following equation:
\begin{equation}
     \mathcal{V} := \mathcal{V}_{1p} \cup \mathcal{V}_{2p} \cup \mathcal{V}_{3p}
\end{equation}
where $\mathcal{V}_{1p}$, $\mathcal{V}_{2p}$ and $\mathcal{V}_{3p}$ respectively represent the nodes associated with three-phase, two-phase and single-phase buses and regulators. With  the inclusion of the voltage regulators as nodes, $\lvert  \mathcal{V}_{3p} \rvert =68$, $\lvert  \mathcal{V}_{2p} \rvert =4$,  $\lvert  \mathcal{V}_{3p} \rvert=56$  which equate to $\lvert  \mathcal{V} \rvert =128$. The set of edges is denoted by $\mathcal{E}$ where $\lvert \mathcal{E} \rvert =127$ and $\mathcal{A} \in \bb R^{\lvert \mathcal{V} \rvert \times \lvert \mathcal{V} \mid}$ is the symmetric adjacency matrix. As $\mathcal{G}$ is a sparse graph, the adjacency matrix is replaced by a coordinate list format (COO) representation denoted by $E_{c} \in \bb R^{2\times 2\lvert \mathcal{E} \rvert }$. $E_{c}$ holds the node pair that is connected by an edge. Owing to the fact that the graph under consideration is undirected in nature, if there is an edge between node pair $(p,q)$ then both $(p,q)$ and $(q,p)$ are included in $E_{c}$.

From this point, we use superscript $k$ to denote the index associated with a data point, subscript $i$ to denote the index of a specific node and subscript $t$ to denote the task index.

The generated dataset contains a feature vector associated with each node in the graph. That is represented by $\vec z^{k}_{i} \in \bb R^{6}$ which is the feature vector for $i^{th}$ node of the $k^{th}$ data point.

Each feature vector holds the value of voltage phasor meaning voltage amplitude ($V_{i}$) and angle ($\phi_{i}$) for three phases. This can be mathematically represented by \eqref{eqn:feature_vec}.
\begin{equation}
\label{eqn:feature_vec}
\vec z_{i}^{k} := [\left.V_i^A, \phi_i^A, V_i^B, \phi_i^B, V_i^C, \phi_i^C\right]^{k}
\end{equation}
where, the superscript $A, B, C$ represent respectively the value associated with Phase-A, Phase-B and Phase-C. It is worth noting that since for the nodes in $ \mathcal{V}_{2p} $ and $\ \mathcal{V}_{1p}$,  values for some specific phases do not exist, they are replaced with zeros.

Stacking all the feature vectors for all $\lvert\mathcal{V}\rvert $ nodes in a graph $\mathcal{G}^{k}$ results in a feature matrix $\mathcal{X}^{k} \in \bb R^{\lvert \mathcal{V} \rvert \times \lvert \mathcal{V} \rvert}$ associated with that graph. A particular row $i$ in the feature matrix $\mathcal{X}^{k}$ corresponds to the feature vector for the $i^{th}$ node. It should be noted the COO representation of the adjacency matrix is the same for all the data points \textit{i.e.}  $E_{c}^{k}=E_{c}, \forall k$. Therefore, the $k^{th}$ input data point of our dataset can be represented by $X^{k} :=(\mathcal{X}^{k},E_{c})$.

For each input data point, there are $5$ labels which are represented respectively by $ y_{\text{detect}}$, $y_{\text{loc}}$, $y_{\text{type}}$, $y_{\text{res}}$, $y_{\text{current}}$. These signify the fault detection label, fault classification label, fault type label, fault resistance label and fault current label. For ease of representation, the subscripts are replaced by the corresponding task index. The first three labels are for classification type prediction and the last two labels are for regression type prediction.
Now we define $Y$ as an ordered list of all the labels which can be represented as follows:-
\begin{equation}
    Y^{k} :=(y^{k}_{1}, y^{k}_{{2}}, y^{k}_{3}, y^{k}_{4}, y^{k}_{5})
\end{equation}
Therefore, our dataset can be succinctly  written as
\begin{equation}
\mathcal{D}^{k}:= \{ (X,Y)^{k} \}
\end{equation}

where a single data point and the corresponding labels are indexed by $k$. The size of the dataset is denoted by $\lvert \mathcal{D}^{k} \rvert = N $ and the training and testing part of the dataset is represented by $\mathcal{D}^{k}_{tr}$ and $\mathcal{D}^{k}_{test}$ respectively, each of which has the size $\lvert \mathcal{D}^{k}_{tr}\rvert =N_{tr}$ and  $ \lvert \mathcal{D}^{k}_{test} \rvert= N_{test}$. Similarly, the inputs associated with the training and testing dataset are expressed as $X_{tr}$ and $X_{test}$ and labels associated are expressed by $Y_{tr}$ and $Y_{test}$.

Now in implementation, we modify the labels of fault current because they can have a varying range of magnitude up to order of $10^3$  or more. This can make the entire optimization process unstable and result in exploding gradients during the training phase. Hence we first normalize the labels to get $\tilde{y}_{t=5}$ followed by taking the negative log which can be expressed as 
\begin{equation} \label{eqn:current_label_transformation}
     y^{k}_{5} := -\ln(\frac{y^{k}_{5}}{\sum^{N}_{k=1} y^{k}_{5}}) 
\end{equation}
This makes the fault current labels in the same range as fault resistance and hence results in much more effective training.

We define our heterogeneous MTL model as $f_{\theta}$,  parameterized by $\theta$. The model can be sectioned into two parts. The first part is the common backbone GNN represented by $g_{\theta^{sh}}$ where $\theta^{sh}$ are the parameters of the common backbone $g$. Now for each task $t$, there is a sequence of separate dense layers which can be represented by $h_{\theta^{t}}$ where $\theta^{t}$ represent parameters associated with a specific task $t$. As there are a total of $T=5$ tasks, we can represent $t$ as $t\in \{1, 2, \cdots, T\}$ which means the network parameters can be expressed as $\theta :=\{\theta^{sh}, \theta^{1}, \theta^{2}, \cdots, \theta^{T} \}$. Now the predicted output $\hat{y}^{k}_{t}$ of a task $t$ for $k^{th}$ data point can be expressed as~\eqref{input-output-relation}
\begin{equation}\label{input-output-relation}
  \hat{y}^{k}_{t} := h_{\theta^{t}}\circ g_{\theta^{sh}}(X^{k}_{tr})         
\end{equation}

For each task, we define a loss function $\mathcal{L}_{t}$ which takes in $X^{k}_{tr}$ and the parameters of the proposed model and computes the loss for that particular task. 
\begin{table}[htbp]
\centering
\caption{Number of parameters for a single batch of size $32$}
\renewcommand{\arraystretch}{0.8}
	\setlength{\tabcolsep}{9pt}
	\resizebox{0.85\linewidth}{!}{
 \setlength{\extrarowheight}{.3em}

\begin{tabular}{lcc}
\toprule
Layers     & $\#$ Parameters & Output Shape \\ \midrule
Input Layer ($\mathcal{H}^{0}$)   & --- & $(32,128,6)$  \\

\underline{\textbf{GNN Backbone($g_{\theta^{sh}}$):}}  &  &\\

GCNConv ($\mathcal{H}^{1}$)   & $224$ & $(32,128,32)$\\

Layer Normalization           & $64$ & $(32,128,32)$ \\

GCNConv ($\mathcal{H}^{2}$)   & $1056$  & $(32,128,32)$\\

Layer Normalization           & $64$  & $(32,128,32)$ \\

GCNConv ($\mathcal{H}^{3}$)   & $1056$  &$(32,128,32)$\\

Concatenation Layer          & --- & $(32,4096)$ \\  

\underline{\textbf{Classification Heads($h_{\theta^{cls}}$):}}      &  & \\

Fault Detection Head ($h_{\theta^{1}}$)   &  $172,974$ & $(32,2)$ \\ 

Fault Localization Head ($h_{\theta^{2}}$)   & $135,328$  & $(32,128)$\\
 
Fault Type Classification Head ($h_{\theta^{3}}$) & $173,222$ & $(32,6)$\\

\underline{\textbf{Regression Heads($h_{\theta^{reg}}$):}}& &\\

Fault Resistance Estimation Head ($h_{\theta^{4}}$)  & $172,969$ & $(32,1)$ \\

Fault Current Estimation Head ($h_{\theta^{5}}$) & $172,969$ & $(32,1)$ \\

\hline

Total        &   $829,926$ & \\

\bottomrule
\end{tabular}}

\label{tab:tp_table}
\end{table}
Each loss is weighted by a weighting factor $w_{t}$. The overall objective function including L2 regularization can be expressed as \eqref{objective_function} where $\lambda \in \bb R$ is the regularization hyperparameter.
\begin{equation}\label{objective_function}
\min_{\theta}  \sum_{k=1}^{N_{tr}} \sum_{t=1}^T w _{t} \mathcal{L}_t ({\theta}, X^{k}_{tr})+\lambda\| \theta \|_{2}
\end{equation}
For the classification layers jointly expressed as $h_{\theta^{cls}}=\{h_{\theta^{1}},h_{\theta^{2}}, h_{\theta^{3}} \}$ the negative log-likelihood (NLL) loss function is used and for the regression layers jointly expressed as $h_{\theta^{reg}}=\{h_{\theta^{4}},h_{\theta^{5}} \}$  the mean squared error (MSE) loss is used. For generality, we express the classification losses as $\mathcal{L}_{cls}$ and regression losses
as $\mathcal{L}_{reg}$ and they are defined as follows over entire training data:-
\begin{equation}
\mathcal{L}_{cls}=-\sum_{k=1}^{N_{tr}} y_{t}^k \log \frac{\exp (h_{\theta^{t}}\circ g_{\theta^{sh}}(X^{k}_{tr})  )}{\sum_{j=1}^m \exp ( h_{\theta^{t}}\circ g_{\theta^{sh}}(X^{k}_{tr}) )}
\end{equation}
where $t \in \{1,2,3\}$ which are the task index for classification task and $m$ is the number of classes for task $t$.
\begin{equation}
    \mathcal{L}_{reg} = \frac{1}{N_{tr}} \sum_{k=1}^{N_{tr}} (\hat{y}^{k}_{t}-y_{t}^{k})^2
\end{equation}
 in this case $t \in \{4,5\}$ which are the indices for regression tasks.
\section{Architecture Detail and Model Training}\label{sec:Architecture_Detail_and_Model_Training}

The GNN backbone of the proposed model consists of $3$ GCN layers which are $\mathcal{H}^{1},\mathcal{H}^{2},\mathcal{H}^{3}$ and $\mathcal{H}^{0}$ is the input layer.

The number of layers is restricted to $3$ to avoid over smoothing issue~\cite{cai2020note} which is a common problem for deep GNNs. For normalization of the features flowing through the layers, layer normalization is used~\cite{ba2016layer}.

In Fig.~\ref{fig:arch} the entire forward propagation through the network is shown. The forward propagation through GNN allows feature representation learning through message passing. In addition, topological information is captured in the learned representations.

After that, learned embedding is extracted from all $128$ nodes which are then flattened and concatenated together to generate a feature vector of dimension $128\times 32 = 4096$. This feature vector is passed through the different heads for different prediction tasks. It is crucial to note that for non-fault events we restrict all the losses except for $\mathcal{L}_{1}$ and  $\mathcal{L}_{3}$, as all the other losses correspond to a fault event. This allows gradient flow (during backpropagation) through the network only for $\mathcal{L}_{1}$ and  $\mathcal{L}_{3}$ loss for non-fault data samples.
\begin{table}[h]
\caption{Hyperparameters and their values}
\centering
\renewcommand{\arraystretch}{0.7}
	\setlength{\tabcolsep}{18pt}
	\resizebox{0.9\linewidth}{!}{
 \setlength{\extrarowheight}{.5em}

    \begin{tabular}{rl}
    \toprule
    \textbf{Hyperparameters} & \textbf{Value} \\
    \midrule
    Batch Size & $32$ 
    \\
    Epochs & $500$
    \\
    Hidden Layer Activation ($g_{\theta^{sh}}$) & ReLU 

    \\ 
    Output Layer Activation ($h_{\theta^{cls}}$) & Log Softmax \\
    
    Output Layer Activation ($h_{\theta^{reg}}$) & -- \\
  
    Optimizer & AdamW \\
    Initial Learning Rate ($\alpha$) & $0.001$ \\ 
    Gradient Clipping Threshold ($\mathcal{C}$)  & $5$ \\
    Weight Decay ($\lambda$) & $10^{-3}$ \\
    Dropout Rate ($D_{r}$) & $0.2$ \\
    Train-Test Split & $80-20$ \\
    $ w_t \rightarrow (w_{1}, w_{2}, w_{3}, w_{4}, w_{5})$ & ($0.01, 0.8, 0.9, 0.1, 0.04, 0.05$) \\
    Random Seed (for reproducibility) & 66
     
    \\
    \bottomrule
    \end{tabular}
    }
\label{tab:parameters_table}
\end{table}

\begin{algorithm}[H]
  \scriptsize
\caption{Training Algorithm of MTL-GNN}
\begin{algorithmic}[1]
\setstretch{1}
\Require $\mathcal{D}_{tr}, \alpha, \lambda, \mathcal{C}, w_{t} $  \Comment{Training set and hyperparameters} 
\Ensure $\theta$ \Comment{Learned parameters of the model}
\State $X_{tr},Y_{tr} \gets \mathcal{D}_{tr} $ 
\While{$ s < \text{number of epochs}$} 
  \For{$k = 1$ to $N_{tr}$}
    \For{$t = 1$ to $T$}

     \State $  \hat{y}^{k}_{t} \gets  h_{\theta^{t}}\circ g_{\theta^{sh}}(X^{k}_{tr})$
\vspace{0.05cm}
   \If{$t = 1$ and $\hat{y}^{k}_{1} == 1$}  \Comment{Non-fault event}
 \vspace{0.2cm}
     \State $\mathcal{L}_s\gets \sum_{k} \sum_{t \in \{1,3 \}} w_{t} \mathcal{L}_{cls}$

  \Else
     \State $\mathcal{L}_s \gets  \sum_{k} \sum_{t} w _{t} \mathcal{L}_t $
  \EndIf
     
     \EndFor
  \EndFor
  \vspace{0.2cm}
   \State $\nabla_\theta \mathcal{L}_s \gets 
\begin{cases}
\nabla_\theta \mathcal{L}_s, & \text{if } \| \nabla_\theta \mathcal{L}_s \| \leq \mathcal{C} \\
\frac{ \mathcal{C}}{\| \nabla_\theta \mathcal{L}_s \|} \cdot \nabla_\theta \mathcal{L}_s, & \text{otherwise}
\end{cases}$
    \vspace{0.2cm}
  \State $\theta_{s} \gets \Call{AdamW}{\theta_{s},\alpha, \lambda, \nabla_\theta \mathcal{L}_s}$ 
 \EndWhile
 \vspace{0.2cm}
\State $ \theta \gets \theta_{s}$

\end{algorithmic}
\label{algo:training}
\end{algorithm}
Without this mechanism, the parameters associated with fault events ($\theta_{2},\theta_{4},\theta_{5}$) would get updated also.

The total number of parameters per layer and output shapes are mentioned in Table~\ref{tab:tp_table}. The model was trained for $500$ epochs with a batch size of 32 with AdamW~\cite{loshchilov2017decoupled} optimizer on an NVIDIA A100
80GB GPU. For stability of training gradient clipping and L2 regularizer are used as well as dropout is used ($D_{r}=0.2$) to prevent overfitting. The hyperparameters are outlined in Table \ref{tab:parameters_table}.

 The training procedure is shown in Algorithm~\ref{algo:training}. For training the model the training dataset $\mathcal{D}_{tr}$ and the hyperparameters : initial learning rate~($\alpha$), regularizer hyperparameter/weight decay~($\lambda$), gradient clipping threshold~($\mathcal{C}$), loss weighting factors~($w_{t}$) need to be specified.
 
 To determine the $w_{t}$ hyperparameters a trial run with the same weight was conducted to get a gauge about which task is easier to learn for the model and after that, these values are set taking into account the importance of the task. We specify the epoch index as $s$ and the corresponding overall weighted loss associated with that epoch is $\mathcal{L}_{s}$. After a forward pass through the network, $\mathcal{L}_{s}$ is calculated followed by the gradient of the loss with respect to model parameters which is $\nabla_\theta\mathcal{L}_{s}$. Gradient clipping is applied if necessary and finally, the AdamW optimizer updates the model parameters based on the defined hyperparameters.

\begin{table*}[!thb]
	\centering
	\renewcommand{\arraystretch}{1}
	\setlength{\tabcolsep}{5pt}
	\caption{Performance of proposed model on all $5$ tasks considering measurement error, resistance change and dataset size }

	\resizebox{0.8\linewidth}{!}{

		\begin{tabular}{l|*{2}{c}|*{4}{c}|*{2}{c}|*{2}{c}|*{2}{c}}
			\toprule
   
			\textbf{Criteria}& 
        \multicolumn{2}{c|}{ \makecell[c]{\bf Fault \\ \bf Detection}} & \multicolumn{4}{c|}{ \makecell[c]{\bf Fault \\ \bf Localization}} &
	\multicolumn{2}{c|}{\makecell[c]{\bf Fault \\ \bf Classification}} & 
        \multicolumn{2}{c|}{\makecell[c]{\bf Fault  Resistance \\ \bf Estimation}}&
        \multicolumn{2}{c}{\makecell[c]{ \bf Fault Current \\ \bf Estimation}}
                \\
                
                \  
 & \multicolumn{2}{c|}{\textbf{-----------------------------------------}}
 & \multicolumn{4}{c|}{\textbf{------------------------------------------}} 
 & \multicolumn{2}{c|}{\makecell[c]{\textbf{---------------------------------------}}} 
 & \multicolumn{2}{c|}{\makecell[c]{\textbf{-----------------------------}}}
 & \multicolumn{2}{c}{\makecell[c]{\textbf{-----------------------------}}}

                \\
			& Balanced Accuracy & F1-Score 
                & $\text{LAR}^{0}$ & $\text{LAR}^{1}$ &  $\text{LAR}^{2}$ &F1-Score
                & \quad Accuracy  & F1-Score 
                & \quad MSE & MAPE  
                & \quad MSE & MAPE 
                \\
   
			\midrule

                \textit{\underline{$\text{Measurement Error}$}} 
                            &   &  
                            &  &  &  & 
                            &  &    
                            &  &    
                            &  &     
                            \\

                $^{\dagger}\text{Low Noise}$: $n \sim \mathcal{N}(0,0.0001)$ 
                            & $1.0$ & $1.0$ 
                            & $0.982$ & $0.999$ & $0.999$ & $0.983$ 
                            & \quad $0.991$ & \quad $0.991$   
                            & \quad $0.108$ & \quad $0.094$   
                            & \quad $0.022$ & \quad $0.008$     
                            \\
                            
			$^{\dagger}\text{Moderate Noise}$: $n \sim \mathcal{N}(0,0.001)$ 
                            & $1.0$ & $1.0$ 
                            & $0.980$ & $0.999$ & $0.999$ & $0.980$
                            & \quad $0.992$ &\quad $0.992$   
                            & \quad $0.111$ & \quad $0.102$   
                            & \quad $0.022$ & \quad $0.008$     
                            \\

                $^{\dagger}\text{High Noise}:$ $n \sim \mathcal{N}(0,0.01)$ 
                            & $1.0$ & $1.0$ 
                            & $0.715$ & $0.910$ & $0.963$ & $0.714$ 
                            & \quad $0.989$ & \quad $0.824$   
                            & \quad $0.208$ & \quad $0.111$   
                            & \quad $0.031$ & \quad $0.010$     
                            \\  
               
                $\text{High Noise}:$ $n \sim \mathcal{N}(0,0.01)$ 
                            & $1.0$ & $1.0$ 
                            & $0.954$ & $0.999$ & $0.999$ & $0.952$ 
                            & \quad $0.992$ & \quad $0.992$   
                            & \quad $0.136$ & \quad $0.116$   
                            & \quad $0.040$ & \quad $0.011$     
                            \\  

            \midrule
            
                \textit{\underline{$\text{Resistance Range Change}$}} 
                            &   &  
                            &  &  &  & 
                            &  &    
                            &  &    
                            &  &     
                            \\

                   \textbf{$0.05 \Omega-20 \Omega$} 
                            & $1.0$ & $1.0$ 
                            & $0.984$ & $0.999$ & $0.999$ & $0.984$
                            & \quad $0.993$ & \quad $0.993$   
                            & \quad $0.133$ & \quad $0.112$   
                            & \quad $0.026$ & \quad $0.008$     
                            \\

			\textbf{$20 \Omega-100 \Omega$} 
                            & $1.0$ & $1.0$ 
                            & $0.983$ & $0.999$ & $0.999$ & $0.982$  
                            & \quad $0.991$ & \quad $0.991$   
                            & \quad $1.468 $ & \quad $0.020$   
                            & \quad $0.008$ & \quad $0.006$     
                            \\

                \textbf{$100 \Omega-500 \Omega$} 
                            & $1.0$ & $1.0$ 
                            & $0.917$ & $0.990$ & $0.999$ & $0.914$  
                            & \quad $0.949$ &\quad  $0.948$   
                            & \quad $68.92$ & \quad $0.021$   
                            & \quad $0.028$ & \quad $0.009$     
                            \\
           \midrule
                                 \textit{\underline{\makecell[l]{\% Samples /Fault Type}}} 
                            &   &  
                            &  &  &  & 
                            &  &    
                            &  &    
                            &  &     
                            \\

                  \textbf{$15\%$} 
                            & $1.0$ & $1.0$ 
                            & $0.934$ & $0.997$ & $0.999$ & $0.931$
                            & \quad $0.991$ & \quad $0.991$   
                            & \quad $0.206$ & \quad $0.302$   
                            & \quad $0.040$ & \quad $0.014$     
                            \\

                   \textbf{$25\%$} 
                            & $1.0$ & $1.0$ 
                            & $0.957$ & $0.995$ & $0.999$ & $0.957$
                            & \quad $0.992$ & \quad $0.992$   
                            & \quad $0.146$ & \quad $0.174$   
                            & \quad $0.030$ & \quad $0.011$     
                            \\

			\textbf{$50\%$} 
                            & $1.0$ & $1.0$ 
                            & $0.975$ & $0.999$ & $0.999$ & $0.975$  
                            & \quad $0.991$ & \quad $0.991$   
                            & \quad $0.211$ & \quad $0.181$   
                            & \quad $0.040$ & \quad $0.011$     
                            \\
 \midrule
                                 \textit{\underline{\makecell[l]{Topology Change}}} 
                            &   &  
                            &  &  &  & 
                            &  &    
                            &  &    
                            &  &     
                            \\

                  $^{\dagger}\text{Open 97-197, Close 151-300}$
                            & $1.0$ & $1.0$ 
                            & $0.988$ & $0.999$ & $0.999$ & $0.987$
                            & \quad $0.993$ & \quad $0.993$ 
                            & \quad $0.091$ & \quad $0.113$ 
                            & \quad $0.024$ & \quad $0.008$ 
                            \\

                   $^{\dagger}\text{Open 18-135, Close 151-300}$
                            & $1.0$ & $1.0$ 
                            & $0.987$ & $0.999$ & $0.999$ & $0.987$
                            & \quad $0.993$ & \quad $0.993$ 
                            & \quad $0.093$ & \quad $0.118$ 
                            & \quad $0.024$ & \quad $0.009$     
                            \\
            \bottomrule
         \multicolumn{12}{l}{$^{\dagger}$Out-of-Distribution (OOD) data } \\

          \bottomrule
		\end{tabular}
	}
 \label{tab:metric_table}
\end{table*}
\section{Numerical Tests}\label{sec:Numerical_Tests}
In this section, we first introduce the evaluation metrics used to assess the model performance, followed by the regression and classification performance of the model. Then we evaluate the model performance on sparse measurements where the sparse node-set is strategically chosen based on an explainability algorithm.
\subsection{Metrics Used for Evaluation of the Model Performance}
For fault detection which is a binary classification task, we report balanced accuracy and f1-score. The reason for reporting balanced accuracy as opposed to accuracy is the class imbalance for fault detection. For fault localization, we report the location accuracy rate (LAR) and f1-score. The $\text{LAR}^{h}$ ($h$ indicate the number of hops) is a common metric to evaluate the performance for fault localization. It quantifies the percentage of correctly identified fault locations within a certain $h$-hop distance from the actual fault location. $\text{LAR}^{0}$ capture the accuracy for exact fault location. In contrast, $\text{LAR}^{1}$ measures the performance of the identified fault locations within $1$-hop distance from the actual fault location. $\text{LAR}^{2}$ computes the same thing but for $2$-hop distance. The reason for computing this metric is to evaluate the model's ability to provide an estimation of the fault's approximate position, even if the model cannot precisely identify the exact fault location. 

For fault type classification we report accuracy and f1-score. In addition, we also use the confusion matrix which provides insight into the model's performance specific to a fault type.

For the regression tasks, fault resistance and fault classification estimation we report MSE and mean absolute percentage error (MAPE) for the test data set. These two metrics are given by the following equation.
\begin{equation}
   \mathrm{MSE}_{test} = \frac{1}{N_{test}} \sum_{k=1}^{N_{test}} (\hat{y}^{k}_{t}-y_{t}^{k})^2 
\end{equation}
\begin{equation}
\mathrm{MAPE}_{test}=\frac{1}{N_{test}} \sum_{k=1}^{N_{test}}\left|\frac{\hat{y}^{k}_t-{y}^{k}_t}{{y}^{k}_t}\right| \times 100
\end{equation}
where $t=4$ indicates the metrics for fault resistance and $t=5$ indicates the metrics for fault current. The reason for reporting MAPE in addition to MSE is the sensitivity to scale for MSE. As we report performance for different ranges of fault current and resistance, this varying range needs to be taken into account. All results are summarized in Table~\ref{tab:metric_table}. For simulating possible measurement errors, we introduce noise $n$ sampled from a zero mean gaussian distribution $\mathcal{N}(0,\sigma_{noise})$ with a variance value set to $\sigma_{noise}$. The variance $\sigma_{noise}$ is respectively set to $0.0001$, $0.001$ and $0.01$ for different levels of noise. We also consider the model performance under varying ranges of fault resistance values which are $0.05 \Omega -20 \Omega$,  $20 \Omega -100 \Omega$ and  $100\Omega-500 \Omega$. Furthermore, we consider the fact there might be a lack of historical data for fault events.  To imitate this, we decrease the size of the dataset and evaluate the model performance on the decreased dataset. The smallest size we consider is $15\%$ of the original which results in only $45$ samples per node. We also report another set of metrics considering two possible topology changes in the feeder system.
\subsection{Performance of the Model for Classification Tasks}
From Table~\ref{tab:metric_table} it is apparent that the model can easily distinguish between a load change event and a fault event which is intuitive given that there are only two classes.
\begin{figure}[htbp]
    \centering
    \includegraphics[width=0.55\linewidth]{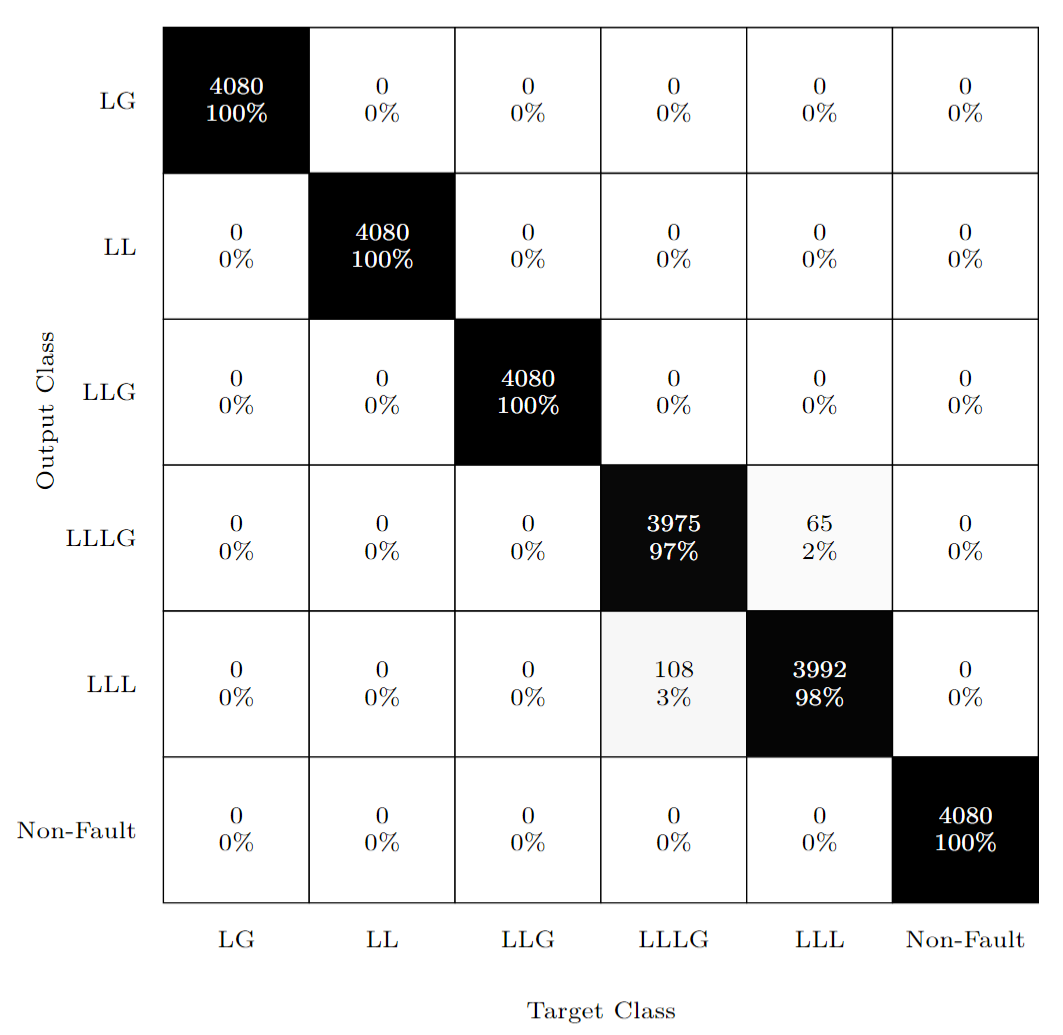}
    \caption{Confusion matrix for fault type classification task. The first three classes are asymmetric faults, the next two classes are symmetric faults and the final class corresponds to non-fault events.}
    \label{fig:confusion_matrix_fault_type}
\end{figure}
In addition, the voltage fluctuation in the case of a fault event differs significantly from that of a load change event. In the case of fault localization, despite the different variations, the model is robust enough to maintain relatively high $\text{LAR}^{1}$ and $\text{LAR}^{2}$. When we consider measurement error, first we evaluate the performance with an out-of-distribution (OOD) test set meaning the training dataset didn't contain noisy samples. For low and moderate noise, even with OOD samples the fault localization performance holds. For high levels of noise, the performance goes down but if the noisy samples are included in the training set the location accuracy improves significantly. However, this robustness to noise can be due to various reasons. Hence, further analysis is performed to evaluate if the robustness holds in case the proposed model is swapped out with other ML models. As regards fault detection, Table~\ref{tab:noise} demonstrates that the measurement noise has no impact on the obtained accuracy. However, for the other $4$ tasks, the robustness to noise does not hold and the performance of other ML models (see Table~\ref{tab:noise}) is significantly worse than the proposed multi-task model as detailed in Table~\ref{tab:metric_table}.  Also, the model manages to sustain relatively high accuracy despite a broad range of resistance values. Similar conclusions can be made for varying dataset sizes. One important thing to note here, even when the model is not able to localize the exact fault point, it can approximate the location with high accuracy.

The performance retains for topology changes in the feeder system. Note that, these metrics are computed with OOD test samples meaning the test samples with the modified topology were never included in the training data. When generating these OOD samples the connectivity information, $E_c$, was changed to reflect the topology change.
\begin{table}[htbp]
\centering
\renewcommand{\arraystretch}{1.3}
\setlength{\tabcolsep}{4pt}
\caption{Performance of traditional ML models under different noise intensity (For resistance range $0.05 \Omega$ - $20 \Omega$)}
\resizebox{0.9\linewidth}{!}{
\begin{tabular}{llccc|cc}
\bottomrule
\textbf{Model} &
  \textbf{\makecell[l]{Noise \\ Level}} &
  \textbf{ \makecell[c]{Detection \\ Accuracy}} &
  \textbf{ \makecell[c]{Classification \\ Accuracy}} &
  \textbf{ \makecell[c]{Localization \\ Accuracy}} &
  \textbf{ \makecell[c]{Resistance \\ Estimation MAPE}} &
  \textbf{\makecell[c]{Current \\Estimation MAPE}} \\ \hline
\multirow{3}{*}{\textbf{\makecell[l]{XGBoost }}} & Low      & $1.0$ & $0.98$ & $0.88$ & $5.27$ & $0.28$ \\
                              & Moderate & $1.0$ & $ 0.97$ & $0.84$ & $6.03$ & $0.43$ \\
                              & High     & $1.0$ & $0.91$ & $0.68$ & $9.07$ & $1.46$ \\ \hline
\multirow{3}{*}{\textbf{\begin{tabular}[c]{@{}l@{}}Random \\ Forest\end{tabular}}} &
  Low &
  $1.0$ &
  $0.98$ &
  $0.90$ &
  $1.15$ &
  $0.07$ \\
                              & Moderate & $1.0$ & $0.97$ & $0.88$ & $1.82$ & $0.11$ \\
                              & High     & $1.0$ & $0.96$ & $0.75$ & $4.38$ & $1.50$ \\
                              \toprule
\end{tabular}
}
\label{tab:noise}
\end{table}
For fault-type classification,  it is important to outline per-class performance as the probability of all fault types is not the same and varies according to fault type; LG (70\% - 80\%), LLG (17\% - 10\%), LL (10\% - 8\%), LLL, LLLG (3\% - 2\%)~\cite{blackburn2006protective}. This means it is more important the model is able to classify the asymmetrical faults compared to the symmetrical faults.

\begin{figure}[htbp]
    \centering
    \includegraphics[width=0.55\linewidth]{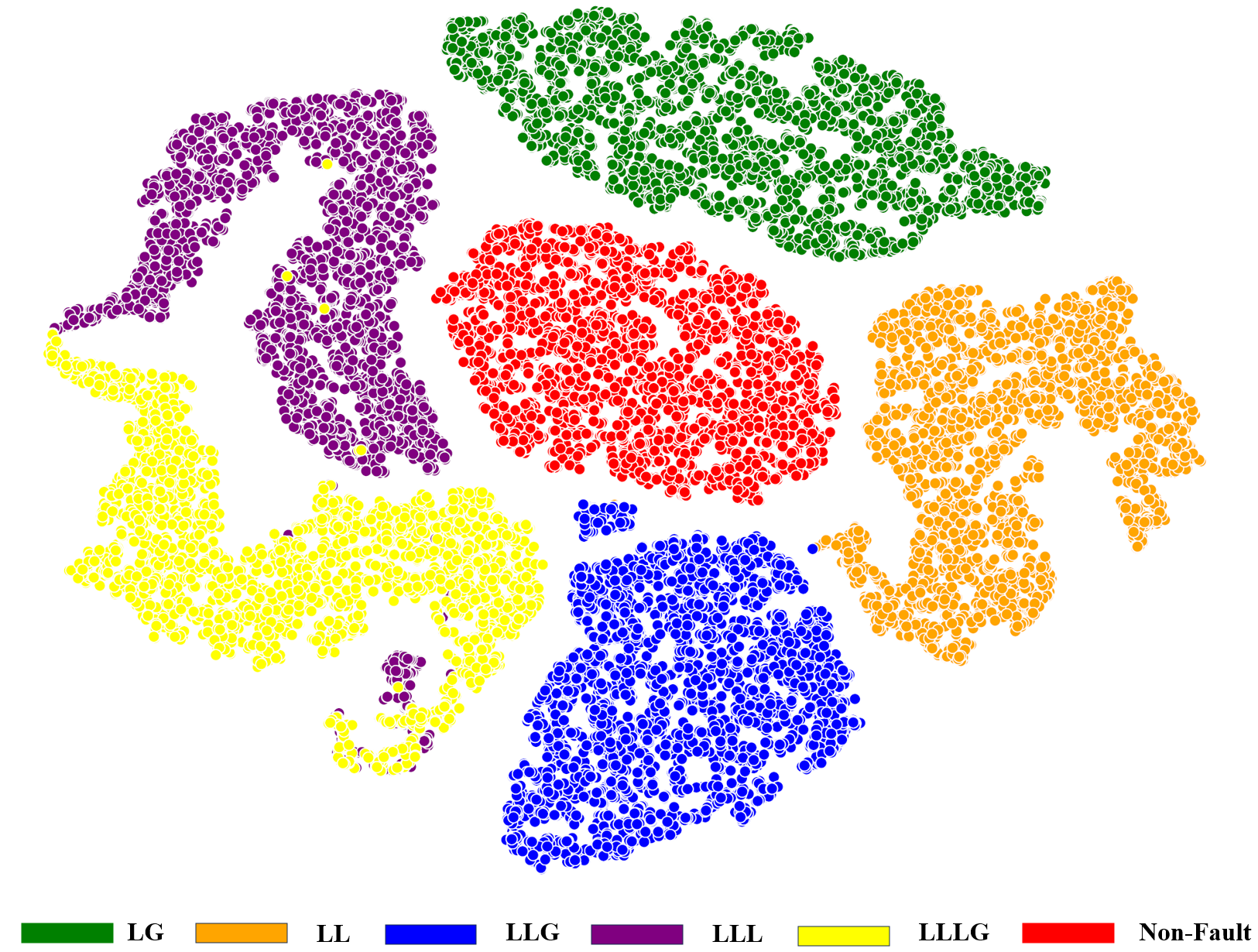}
    \caption{t-SNE visualization of last layer features of fault type classification head $h_{\theta}^{3}$ for the test set $\mathcal{D}_{test}$. Except for a few samples in the LLL fault type and LLLG fault type, most features are clearly separable based on fault type.}
    \label{fig:tsne_fautlt_type_last_layer_features}
\end{figure}

 The confusion matrix shown in Fig.~\ref{fig:confusion_matrix_fault_type} summarizes the performance for each fault type. The misclassified fault types belong to the LLL and LLLG fault class which is further corroborated by the t-SNE visualization of the last layer features of $h_{\theta}^{3}$ shown in Fig.~\ref{fig:tsne_fautlt_type_last_layer_features}. Samples from other fault types are clearly separable even when they are projected into a 2-dimensional feature space.

\subsection{Performance of the Model for Regression Tasks}
For both regression tasks, the performance of the model remains consistently high across all the variations considered. Even though for resistance levels $100\Omega - 500\Omega$, MSE is high due to the scale range, MAPE remains low similar to other resistance levels.

\begin{figure}[htbp]
    \centering
\includegraphics[width=0.55\linewidth]{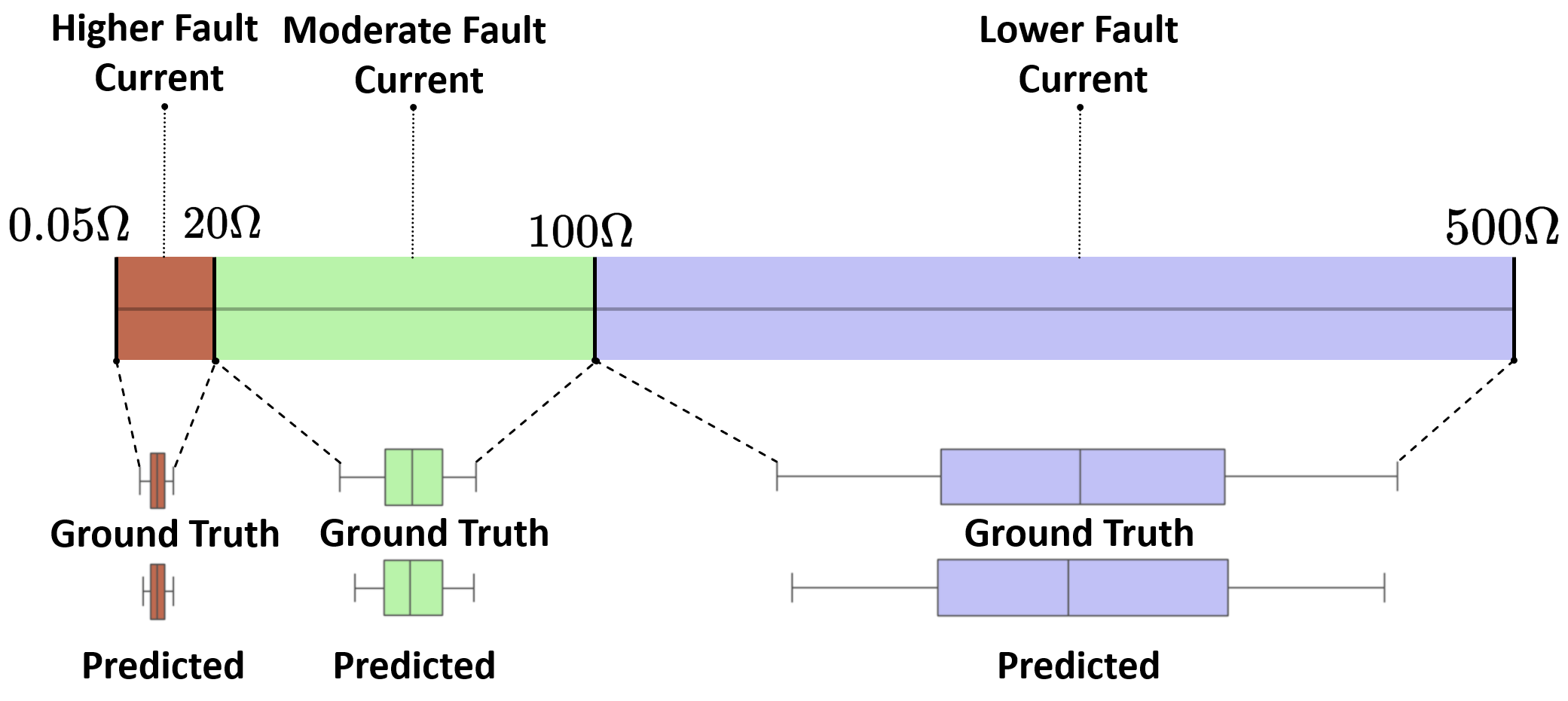}
    \caption{Boxplot of both the ground truth and predicted distribution of the fault resistance.}
    \label{fig:box-plot}
\end{figure}

Fig.~\ref{fig:box-plot} visualizes the range and variance difference among the three fault resistance levels considered in this paper. For resistance level $0.05\Omega-20\Omega$, the fault resistance variance is the lowest, but the fault current level is higher compared to other resistance levels. For resistance level $100 \Omega-500\Omega$, it is exactly the opposite. This further highlights the scale sensitivity issue with MSE that results in a high MSE score with the resistance level $100\Omega-500\Omega$. As MAPE is computed as a percentage of the actual value, it is scale-independent and is preferred with metrics that involve a broader range and large variance which is the case for $100\Omega-500\Omega$.

\begin{figure}[htbp]
    \centering
    \includegraphics[width=0.6\linewidth]{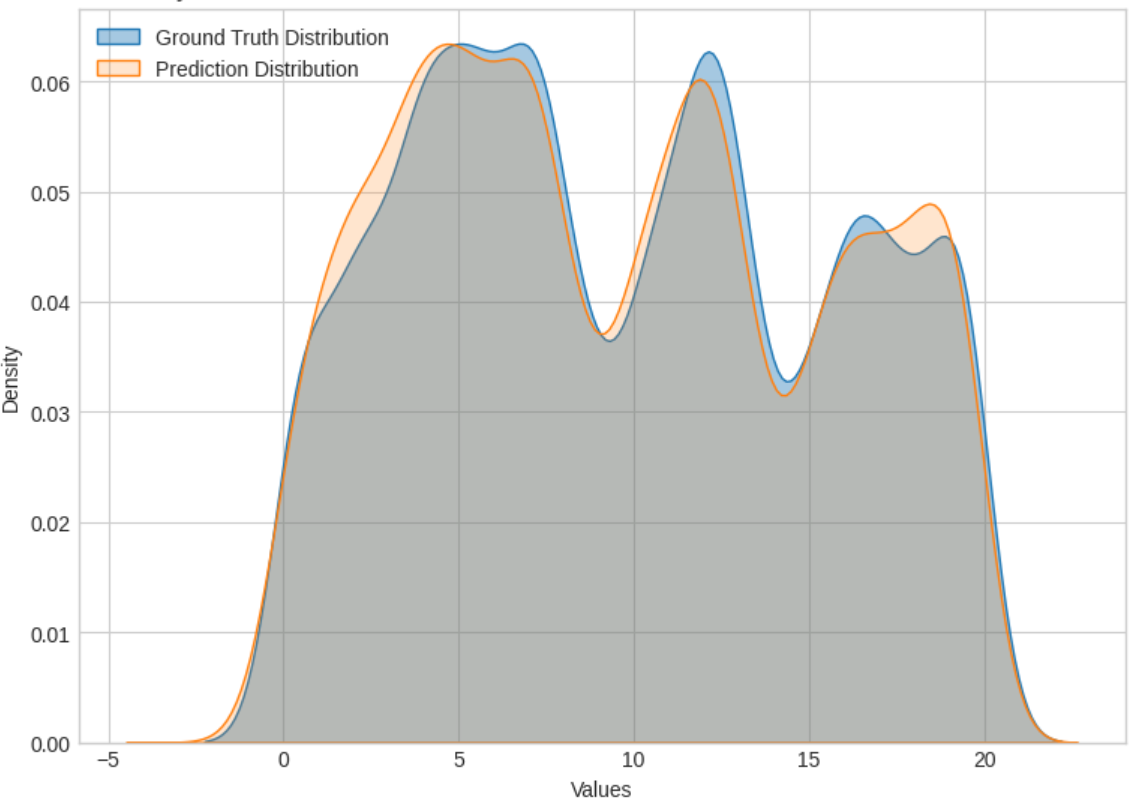}
    \caption{Ground truth ($y_{4}$) and predicted ($\hat{y}_{4}$) fault resistance distribution on the test set ($\mathcal{D}_{test}$). The grey section is the correctly predicted part of the distribution.}
    \label{fig:res_distribution}
\end{figure}

\begin{figure}[htbp]
    \centering
    \includegraphics[width=0.6\linewidth]{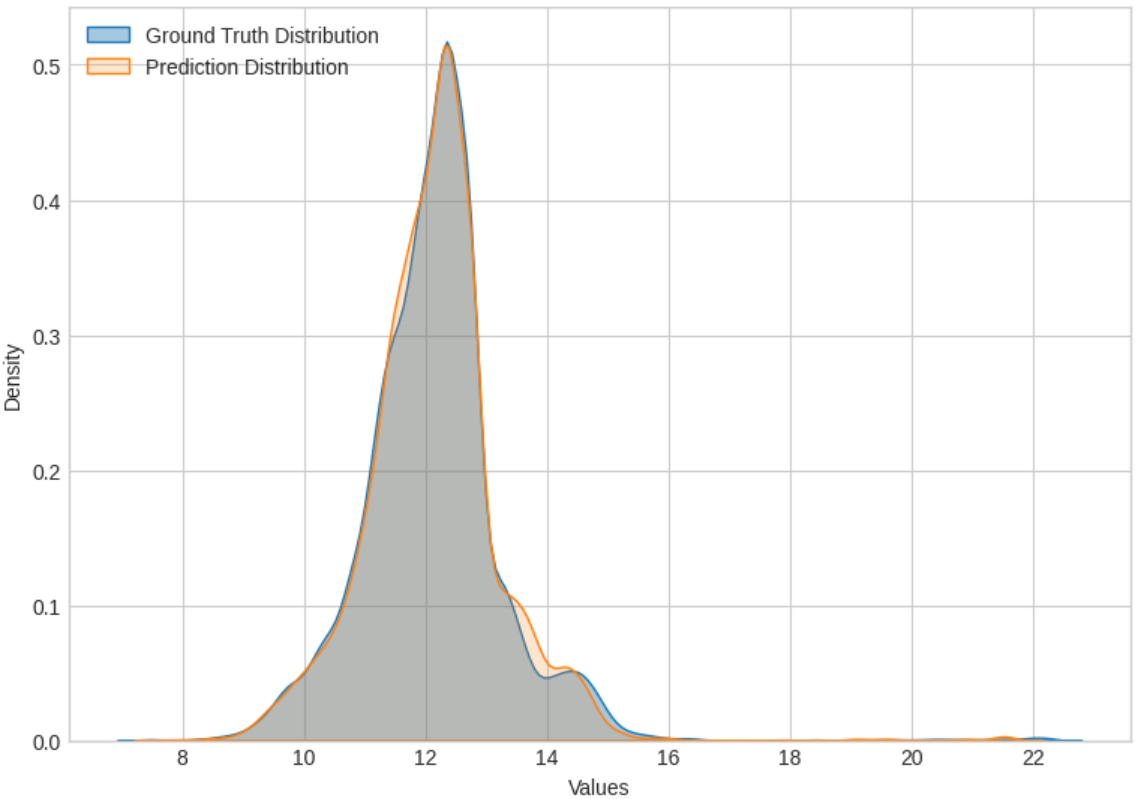}
    \caption{Ground truth ($y_{5}$) and predicted ($\hat{y}_{5}$)  redefined fault current distribution on the test set ($\mathcal{D}_{test}$). The grey section is the correctly predicted part of the distribution.}
    \label{fig:current_distribution}
\end{figure}

Fig.~\ref{fig:res_distribution} shows the distribution plot for the ground truth and predicted value for fault resistance on the test set. For the most part, the predicted distribution closely resembles the ground truth distribution. Similarly, Fig.~\ref{fig:current_distribution} shows the ground truth distribution and predicted distribution for the fault current. One important thing to mention here, this plot is for the transformed fault current label mentioned in \eqref{eqn:current_label_transformation} rather than the actual fault current.

To get the actual fault current prediction and ground truth labels, the inverse operation of \eqref{eqn:current_label_transformation} can be performed which is given by the following equation:
\begin{equation}
      y^{k}_{5} := {\sum^{N}_{k=1} y^{k}_{5}} \times \exp (-y^{k}_{5})
\end{equation}

\subsection{Using Explainability to Identify Key Nodes}
Out of all the tasks, fault localization is the most important given that the model performs really well for fault detection. Considering practical implications, for the remaining three tasks, it is acceptable if the model can approximate the prediction.
So far in analysis, we assumed complete observability which implies all node voltage phasors can be measured. As the scale of the distribution system grows it becomes more and more harder to maintain complete observability considering cost and system complexity. Therefore, we propose a novel approach to locate key nodes using an explainability algorithm GNNExplainer proposed by~\cite{ying2019gnnexplainer}. GNNExplainer uses an optimization that maximizes the mutual information between a GNN’s prediction and distribution of possible subgraph structures to identify important subgraphs.
The most notable characteristic of this algorithm is it does not require ground truth labels. Using this algorithm we generate four sparse node sets $\mathcal{V}_{10\%}$, $\mathcal{V}_{20\%}$, $\mathcal{V}_{50\%}$ and $\mathcal{V}_{75\%}$ where the subscript denotes the percentage of nodes out of $128$ which has data available. The feature vector of the rest of the nodes is set to $0$.

\begin{algorithm}[htbp]
  \scriptsize
\caption{ Sparse Node-Set Generation Algorithm}
\begin{algorithmic}[1]
\setstretch{1}
\Require $X_{test}, epoch_{GE}, h_{\theta^{2}}\circ g_{\theta^{sh}} (trained) , w_{th}$ 
\Ensure $\mathcal{V}_{x\%}$
 \For{$k = 1$ to $N_{test}$}
 \State  $E^{k}_{\mathcal{I}} \gets \Call{GNNExplainer}{X^{k}_{test}, h_{\theta^{2}}\circ g_{\theta^{sh}}, epoch_{GE}}$
  \For{$p = 1$ to $\mid \mathcal{E} \mid$}
   \If{$E^{k}_{\mathcal{I}}(p)>w_{th}$}  \Comment{Thresholding with $w_{th}$}
   \vspace{0.05cm}
    \State $E_{w_{th}}^{k}(p) \gets$ 1
    \Else
     \State $E_{w_{th}}^{k}(p) \gets$ 0
   \EndIf
 \EndFor
 \vspace{0.05cm}
 \State $ \{ \mathcal{V}_{sparse} \}^{k} \gets \Call{GetConnectedNodePairs}{E^{k}_{w_{th}}}$
 \State $\mathcal{V}_{x\%} \gets \bigcup_{k}  \{ \mathcal{V}_{sparse} \}^{k} $
\EndFor
\end{algorithmic}
\label{algo:sparse_node_set}
\end{algorithm}

To generate these sets we pass each test data sample $X_{test}^{k}$ and the trained model with localization head $h_{\theta^{2}}\circ g_{\theta^{sh}}$ to the GNNExplainer algorithm which generates an edge importance vector given by $E_{\mathcal{I}} \in \bb R^{\mid \mathcal{E} \mid }$, indexed by $p$. For each edge in $E_{c}$ a corresponding weight is generated which signifies the importance of that edge. A weight value closer to $1$ means a more important edge and a value closer to $0$ means a less important edge. We threshold the values in  $E_{\mathcal{I}}$ with a threshold $w_{th}$. After thresholding, the transformed edge importance vector is expressed by $E_{w_{th}} \in \bb R^{\mid \mathcal{E} \mid}$. Edges that have an importance score of more than $w_{th}$ are kept and the rest are disregarded. The nodes connected to the edges in $E_{w_{th}}$ after thresholding are regarded as the important nodes for the $k^{th}$ data point. In this way, for each data point, we get a sparse node set given by $ \{ \mathcal{V}_{sparse} \}^{k}$.  Then the union of these sparse important node sets generates the final sparse node set which is given by the following equation:-

\begin{equation}
      \mathcal{V}_{x\%} := \bigcup_{k=1}^{N}  \{ \mathcal{V}_{sparse} \}^{k}
\end{equation}

where the $x\%$ value depends on the threshold value. We set the value of threshold $w_{th}$ to respectively $0.57, 0.52343$, $0.487$ and $0.44$ to generate$\mathcal{V}_{10\%}$, $\mathcal{V}_{20\%}$,     $\mathcal{V}_{50\%}$ and $\mathcal{V}_{75\%}$.

The entire process is described in the Algorithm~\ref{algo:sparse_node_set}. In addition to other parameters mentioned above, an epoch number needs to be specified for the internal optimization of the GNNExplainer.

\begin{table}[htbp]
\centering
\caption{ $\text{LAR}^{h}$ and F1-Score with Sparse Node-Set}
\renewcommand{\arraystretch}{1.1}
	\setlength{\tabcolsep}{9pt}
	\resizebox{0.6\linewidth}{!}{
 \setlength{\extrarowheight}{.3em}

\begin{tabular}{lcccc}
\toprule
Sparse Sets     & $\text{LAR}^{0}$& $\text{LAR}^{1}$ & $\text{LAR}^{2}$ & F1-Score  \\ \midrule
\textbf{$\mathcal{V}_{75\%}$}  & 0.975 & 0.999 & 0.999 & 0.975  \\
\textbf{$\mathcal{V}_{50\%}$}  & \underline{0.962} & 0.998 & 0.999 & \underline{0.962}  \\ 

\textbf{$\mathcal{V}_{20\%}$} & \underline{0.942} & \underline{0.987} & \underline{0.992} & \underline{0.943} \\

\textbf{$\mathcal{V}_{10\%}$} & 0.849 & 0.932 & 0.955 & 0.849\\

\textbf{$\mathcal{V}^{random_{avg}}_{50\%}$}  & \underline{0.916} & 0.993 & 0.999 & \underline{0.913}  \\
\textbf{$\mathcal{V}^{random_{min}}_{50\%}$}  & \underline{0.886} & 0.989 & 0.998 & \underline{0.881}  \\

\bottomrule
\end{tabular}}

\label{tab:results_sparse}
\end{table}

To validate this approach we also randomly sample $50\%$ nodes several times and train the model to generate fault localization metrics on the test set. $\mathcal{V}_{50\%}^{random_{avg}}$ represent the average of these metrics and $\mathcal{V}_{50\%}^{random_{min}}$ represent the minimum of these metrics from the samples generated.
The results of this analysis are summarized in Table~\ref{tab:results_sparse}. It is apparent that the model is robust enough to maintain the $\text{LAR}^{1}$ and $\text{LAR}^{2}$ irrespective of which sparse node-set is used. But for $\text{LAR}^{0}$ and $\text{F1-Score}$, the sparse set generated with GNNExplainer is comparatively better. $\mathcal{V}_{50\%}$  has a $4.78\%$ increase in  $\text{LAR}^{0}$ compared to $\mathcal{V}^{random_{avg}}_{50\%}$ and a $7.9\%$ increase compared to $\mathcal{V}^{random_{min}}_{50\%}$ which means the sparse node-set generation algorithm was able to identify the important node set for fault localization. 
For $\mathcal{V}_{10\%}$ which represents only $10\%$ the nodes, even though $\text{LAR}^{0}$ goes down significantly the $\text{LAR}^{1}$ and $\text{LAR}^{2}$ still holds up. Note that, state estimation methods can be used to approximate the measurements from other nodes. To this end, we outline the benefits gained by identifying important nodes through GNNExplainer:

\begin{enumerate}
    \item\textbf{Reduction in cost:} There are several data acquisition systems in a distribution network which include phasor measurement unit (PMU), micro-phasor measurement unit ($\mu$-PMU), distribution-level PMU (D-PMU) and smart meters. As the scale of the distribution system increases, the number of measurements increases. This limits a complete observability of the system while keeping the cost of the system within bounds. Identifying the important nodes and collecting data only from those nodes will significantly decrease the overall cost.

    \item \textbf{Reduction in system complexity:} With a large number of measurments the system complexity increases substantially as time synchronization and latency in data processing come into the picture. An alternative way to maintain the observability of the whole system with fewer data acquisition units is to use state estimation techniques. However, even with state estimation methods there is the possibility of running into convergence issues and accumulation of errors. These issues can be avoided or their effect can be subsided if the system is designed with a lower number of nodes through the proposed approach.

    \item \textbf{Reduction in training overhead:} As the scale of the distribution system grows, the training overhead of the proposed model will also increase. Generating sparse features by setting unmeasured values to zero, will greatly reduce the training overhead.

    \item \textbf{No need for ground truth labels:} GNNExplainer does not require any ground truth labels for the important nodes. This means no domain or expert knowledge about the distribution system is required to annotate important nodes in advance as the algorithm can select those.
\end{enumerate}

\section{Considerations for Implementation in a Practical Setting}
The proposed method deals with the challenges that come with a practical deployment. However, there are some practical considerations that need to be taken into account including thorough evaluation of the system before deployment, alternative software choice for simulation and day-to-day operation strategy for grid operators. In this section, we discuss how to address these, along with citing relevant research that offers insights into how the proposed method will work in practice.

The National Renewable Energy Laboratory (NREL) has developed a repository with synthetic but realistic data that contains feeder systems with millions of buses and also provides OpenDSS scripts for these distribution systems. This repository named SMART-DS~\cite{palmintier2020smart} can be used to test the proposed method under different realistic conditions. Another option is to use a framework like OPAL-RT that provides a distribution management system (DMS) that is capable of real-time simulation and provides compatibility with OpenDSS. 

A notable work that uses D-PMU data simulated with OPAL-RT in a physical test-bed to detect events is \cite{stifter2018real}. The framework provided in this work follows IEEE standards throughout and can be used to validate the performance of the proposed method in a realistic setting. Another work that has a working prototype of using $\mu$-PMUs to diagnose distribution level events is \cite{liao2016micro} which is a distribution network at Lawrence Berkeley National Laboratory. This work is a proof-of-concept for how our proposed method can be used in real time with data aggregated from $\mu$-PMUs.

We have investigated the changes in the observability of nodes.  For example, an observability of approximately $20\%$ of the nodes still provides a high accuracy in predictions as shown in Table~\ref{tab:results_sparse}. 
It is worth noting that state estimation methods can be used to get observability of these $20\%$ nodes which has been explored in previous works~\cite{pignati2016fault,malandra2018impact}. Obviously using state estimation methods will introduce noise into observations but as shown in Table~\ref{tab:metric_table} our proposed method is robust to noise.
\begin{figure*}[!t]
    \centering
\includegraphics[width=0.62\linewidth]{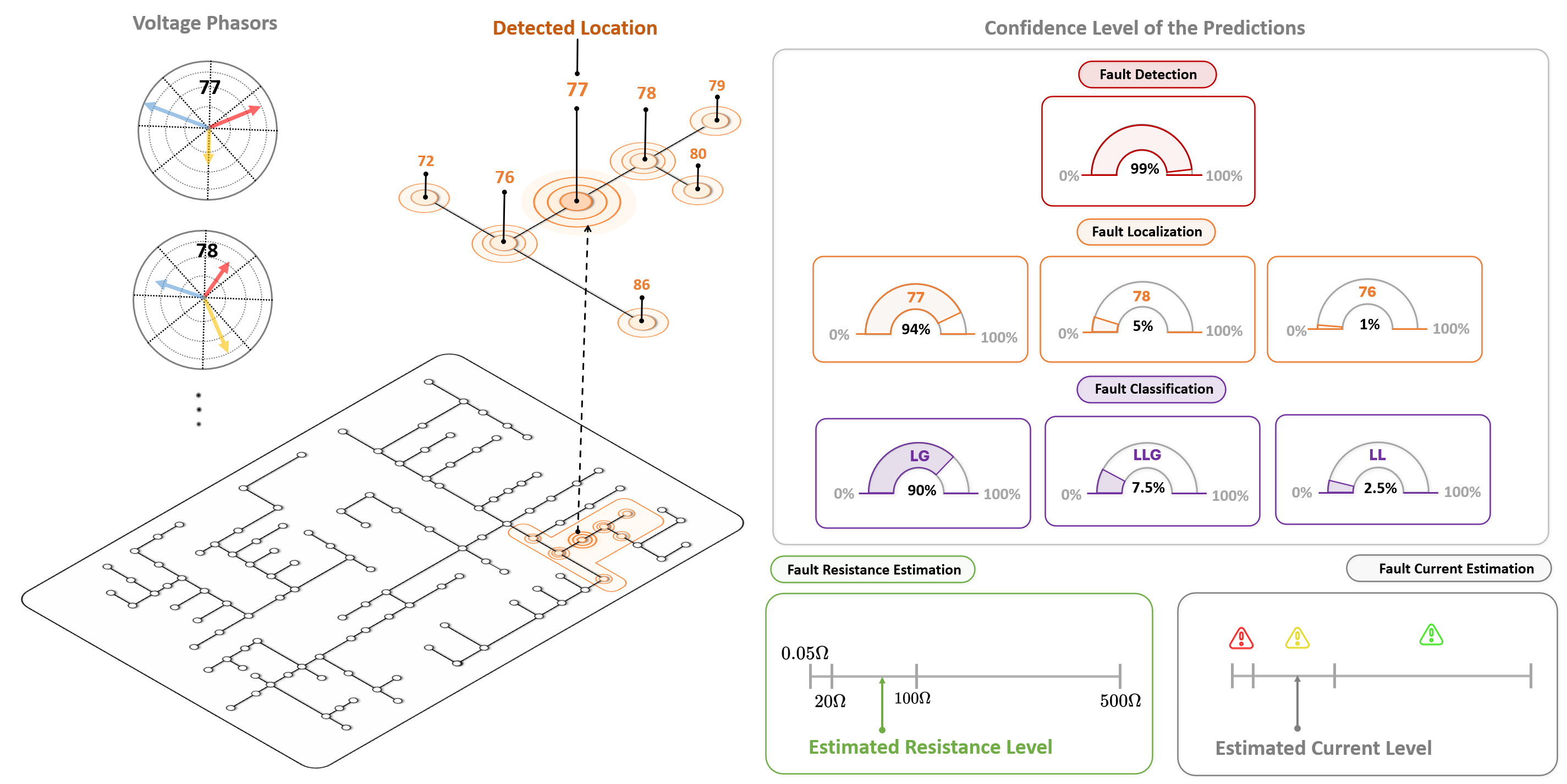}
\caption{Envisioned user interface for grid operators. (\textbf{Left}) Detected fault location (in this example node $77$) in the distribution grid along with 1-hop nodes (in this example 76, 78) and 2-hop nodes (in this example 72, 86, 79, 80) in the neighborhood of the predicted node.
 (\textbf{Right}) The confidence level behind the predictions provides grid operators with additional information to verify the reliability of the prediction and fault resistance and fault current estimation allow grid operators to take appropriate safety measures for fault isolation and clearance.}
    \label{fig:ui_viz}
\end{figure*}
Another practical aspect to consider is how the grid operators will manage the system on a day-to-day basis. To ensure the reliability of model predictions, grid operators need access to more information. Fig.~\ref{fig:ui_viz} illustrates how we envision a user interface. In this figure, a sample data point with fault event at node $77$ is used for the visualization. The grid operators will be provided by both the predicted location and the neighboring nodes within 1-hop and 2-hop distances. Based on the results in Table~\ref{tab:metric_table} almost $100\%$ of the time the predicted fault location is in the 2-hop neighborhood of the actual faulty node. In addition, grid operators will have access to the voltage phasors of these nodes which provides more context behind the predictions. Furthermore, the confidence level of these predictions will allow grid operators to make an informed decision and undertake necessary safety measures before taking any action.

\section{Conclusion}\label{sec:Conclusion_and_Scope_for_Future_Work}

In this paper, we proposed an MTL-GNN capable of performing $5$ different tasks simultaneously even with a sparse node-set. This sparse node-set is generated with a novel algorithm and to the best of our knowledge this is the first work that uses a GNN explainability algorithm for an informed node selection. There are some challenges associated with the proposed method that can be addressed in an extension of this work. The proposed architecture requires more hyperparameter tuning compared to a single-task learning model. A possible future work is automating the process of hyperparameter selections. 

Besides that, the number of tasks needed for a distribution network needs to be specified in advance to configure the model architecture. For example, if the power distribution system requires only fault detection and localization, the architecture needs to be modified prior to deployment. A more adaptive framework for optional number of tasks can be of interest.

\bibliography{ref.bib}

\begin{thebibliography}{10}
\providecommand{\url}[1]{#1}
\csname url@samestyle\endcsname
\providecommand{\newblock}{\relax}
\providecommand{\bibinfo}[2]{#2}
\providecommand{\BIBentrySTDinterwordspacing}{\spaceskip=0pt\relax}
\providecommand{\BIBentryALTinterwordstretchfactor}{4}
\providecommand{\BIBentryALTinterwordspacing}{\spaceskip=\fontdimen2\font plus
\BIBentryALTinterwordstretchfactor\fontdimen3\font minus \fontdimen4\font\relax}
\providecommand{\BIBforeignlanguage}[2]{{%
\expandafter\ifx\csname l@#1\endcsname\relax
\typeout{** WARNING: IEEEtran.bst: No hyphenation pattern has been}%
\typeout{** loaded for the language `#1'. Using the pattern for}%
\typeout{** the default language instead.}%
\else
\language=\csname l@#1\endcsname
\fi
#2}}
\providecommand{\BIBdecl}{\relax}
\BIBdecl

\bibitem{veloza2016analysis}
O.~P. Veloza and F.~Santamaria, ``Analysis of major blackouts from 2003 to 2015: {Classification} of incidents and review of main causes,'' \emph{The Electricity Journal}, vol.~29, no.~7, pp. 42--49, 2016.

\bibitem{guo2017critical}
H.~Guo, C.~Zheng, H.~H.-C. Iu, and T.~Fernando, ``A critical review of cascading failure analysis and modeling of power system,'' \emph{Renewable and Sustainable Energy Reviews}, vol.~80, pp. 9--22, 2017.

\bibitem{blackburn2006protective}
J.~L. Blackburn and T.~J. Domin, \emph{Protective relaying: principles and applications}.\hskip 1em plus 0.5em minus 0.4em\relax CRC press, 2006.

\bibitem{bahmanyar2017comparison}
A.~Bahmanyar, S.~Jamali, A.~Estebsari, and E.~Bompard, ``A comparison framework for distribution system outage and fault location methods,'' \emph{Electric Power Systems Research}, vol. 145, pp. 19--34, 2017.

\bibitem{dashti2021survey}
R.~Dashti, M.~Daisy, H.~Mirshekali, H.~R. Shaker, and M.~H. Aliabadi, ``A survey of fault prediction and location methods in electrical energy distribution networks,'' \emph{Measurement}, vol. 184, p. 109947, 2021.

\bibitem{de2023fault}
J.~De~La~Cruz, E.~G{\'o}mez-Luna, M.~Ali, J.~C. Vasquez, and J.~M. Guerrero, ``Fault location for distribution smart grids: Literature overview, challenges, solutions, and future trends,'' \emph{Energies}, vol.~16, no.~5, p. 2280, 2023.

\bibitem{javadian2009determining}
S.~Javadian, A.~Nasrabadi, M.-R. Haghifam, and J.~Rezvantalab, ``Determining fault's type and accurate location in distribution systems with {DG} using {MLP} neural networks,'' in \emph{International conference on clean electrical power}.\hskip 1em plus 0.5em minus 0.4em\relax IEEE, 2009, pp. 284--289.

\bibitem{tokel2018new}
H.~A. Tokel, R.~Al~Halaseh, G.~Alirezaei, and R.~Mathar, ``A new approach for machine learning-based fault detection and classification in power systems,'' in \emph{IEEE Power \& Energy Society Innovative Smart Grid Technologies Conference (ISGT)}, 2018, pp. 1--5.

\bibitem{guo2017deep}
M.-F. Guo, X.-D. Zeng, D.-Y. Chen, and N.-C. Yang, ``Deep-learning-based earth fault detection using continuous wavelet transform and convolutional neural network in resonant grounding distribution systems,'' \emph{IEEE Sensors Journal}, vol.~18, no.~3, pp. 1291--1300, 2017.

\bibitem{li2019real}
W.~Li, D.~Deka, M.~Chertkov, and M.~Wang, ``Real-time faulted line localization and {PMU} placement in power systems through convolutional neural networks,'' \emph{IEEE Transactions on Power Systems}, vol.~34, no.~6, pp. 4640--4651, 2019.

\bibitem{zou2022double}
M.~Zou, Y.~Zhao, D.~Yan, X.~Tang, P.~Duan, and S.~Liu, ``Double convolutional neural network for fault identification of power distribution network,'' \emph{Electric Power Systems Research}, vol. 210, p. 108085, 2022.

\bibitem{souhe2022fault}
F.~G.~Y. Souhe, A.~T. Boum, P.~Ele, C.~F. Mbey, V.~J.~F. Kakeu \emph{et~al.}, ``Fault detection, classification and location in power distribution smart grid using smart meters data,'' \emph{Journal of Applied Science and Engineering}, vol.~26, no.~1, pp. 23--34, 2022.

\bibitem{paul2022knowledge}
S.~Paul, S.~Grijalva, M.~J. Aparicio, and M.~J. Reno, ``Knowledge-based fault diagnosis for a distribution system with high pv penetration,'' in \emph{IEEE Power \& Energy Society Innovative Smart Grid Technologies Conference (ISGT)}, 2022, pp. 1--5.

\bibitem{shadi2022real}
M.~R. Shadi, M.-T. Ameli, and S.~Azad, ``A real-time hierarchical framework for fault detection, classification, and location in power systems using {PMU}s data and deep learning,'' \emph{International Journal of Electrical Power \& Energy Systems}, vol. 134, p. 107399, 2022.

\bibitem{chen2019fault}
K.~Chen, J.~Hu, Y.~Zhang, Z.~Yu, and J.~He, ``Fault location in power distribution systems via deep graph convolutional networks,'' \emph{IEEE Journal on Selected Areas in Communications}, vol.~38, no.~1, pp. 119--131, 2019.

\bibitem{sun2021distribution}
H.~Sun, S.~Kawano, D.~Nikovski, T.~Takano, and K.~Mori, ``Distribution fault location using graph neural network with both node and link attributes,'' in \emph{IEEE PES Innovative Smart Grid Technologies Europe (ISGT Europe)}, 2021, pp. 1--6.

\bibitem{de2021fault}
J.~T. de~Freitas and F.~G.~F. Coelho, ``Fault localization method for power distribution systems based on gated graph neural networks,'' \emph{Electrical Engineering}, vol. 103, no.~5, pp. 2259--2266, 2021.

\bibitem{li2021ppgn}
W.~Li and D.~Deka, ``{PPGN}: Physics-preserved graph networks for real-time fault location in distribution systems with limited observation and labels,'' \emph{arXiv preprint arXiv:2107.02275}, 2021.

\bibitem{mo2022sr}
H.~Mo, Y.~Peng, W.~Wei, W.~Xi, and T.~Cai, ``{SR-GNN} based fault classification and location in power distribution network,'' \emph{Energies}, vol.~16, no.~1, p. 433, 2022.

\bibitem{hu2022fault}
J.~Hu, W.~Hu, J.~Chen, D.~Cao, Z.~Zhang, Z.~Liu, Z.~Chen, and F.~Blaabjerg, ``Fault location and classification for distribution systems based on deep graph learning methods,'' \emph{Journal of Modern Power Systems and Clean Energy}, vol.~11, no.~1, pp. 35--51, 2022.

\bibitem{nguyen2023spatial}
B.~L. Nguyen, T.~V. Vu, T.-T. Nguyen, M.~Panwar, and R.~Hovsapian, ``Spatial-temporal recurrent graph neural networks for fault diagnostics in power distribution systems,'' \emph{IEEE Access}, 2023.

\bibitem{liang2015fault}
R.~Liang, G.~Fu, X.~Zhu, and X.~Xue, ``Fault location based on single terminal travelling wave analysis in radial distribution network,'' \emph{International Journal of Electrical Power \& Energy Systems}, vol.~66, pp. 160--165, 2015.

\bibitem{shi2018travelling}
S.~Shi, A.~Lei, X.~He, S.~Mirsaeidi, and X.~Dong, ``Travelling waves-based fault location scheme for feeders in power distribution network,'' \emph{The Journal of Engineering}, vol. 2018, no.~15, pp. 1326--1329, 2018.

\bibitem{wang2020traveling}
Y.~Wang, T.~Zheng, C.~Yang, and L.~Yu, ``Traveling-wave based fault location for phase-to-ground fault in non-effectively earthed distribution networks,'' \emph{Energies}, vol.~13, no.~19, p. 5028, 2020.

\bibitem{tashakkori2019fault}
A.~Tashakkori, P.~J. Wolfs, S.~Islam, and A.~Abu-Siada, ``Fault location on radial distribution networks via distributed synchronized traveling wave detectors,'' \emph{IEEE Transactions on Power Delivery}, vol.~35, no.~3, pp. 1553--1562, 2019.

\bibitem{krishnathevar2011generalized}
R.~Krishnathevar and E.~E. Ngu, ``Generalized impedance-based fault location for distribution systems,'' \emph{IEEE transactions on power delivery}, vol.~27, no.~1, pp. 449--451, 2011.

\bibitem{das2012distribution}
S.~Das, N.~Karnik, and S.~Santoso, ``Distribution fault-locating algorithms using current only,'' \emph{IEEE transactions on power delivery}, vol.~27, no.~3, pp. 1144--1153, 2012.

\bibitem{dashti2014accuracy}
R.~Dashti and J.~Sadeh, ``Accuracy improvement of impedance-based fault location method for power distribution network using distributed-parameter line model,'' \emph{International Transactions on Electrical Energy Systems}, vol.~24, no.~3, pp. 318--334, 2014.

\bibitem{jia2016high}
K.~Jia, T.~Bi, Z.~Ren, D.~W. Thomas, and M.~Sumner, ``High frequency impedance based fault location in distribution system with {DGs},'' \emph{IEEE Transactions on Smart Grid}, vol.~9, no.~2, pp. 807--816, 2016.

\bibitem{gautam2012detection}
S.~Gautam and S.~M. Brahma, ``Detection of high impedance fault in power distribution systems using mathematical morphology,'' \emph{IEEE Transactions on Power Systems}, vol.~28, no.~2, pp. 1226--1234, 2012.

\bibitem{sekar2017combined}
K.~Sekar and N.~K. Mohanty, ``Combined mathematical morphology and data mining based high impedance fault detection,'' \emph{Energy Procedia}, vol. 117, pp. 417--423, 2017.

\bibitem{gush2018fault}
T.~Gush, S.~B.~A. Bukhari, R.~Haider, S.~Admasie, Y.-S. Oh, G.-J. Cho, and C.-H. Kim, ``Fault detection and location in a microgrid using mathematical morphology and recursive least square methods,'' \emph{International Journal of Electrical Power \& Energy Systems}, vol. 102, pp. 324--331, 2018.

\bibitem{bayati2021mathematical}
N.~Bayati, H.~R. Baghaee, A.~Hajizadeh, M.~Soltani, and Z.~Lin, ``Mathematical morphology-based local fault detection in dc microgrid clusters,'' \emph{Electric Power Systems Research}, vol. 192, p. 106981, 2021.

\bibitem{wilches2022algorithm}
F.~Wilches-Bernal, M.~Jim{\'e}nez-Aparicio, and M.~J. Reno, ``An algorithm for fast fault location and classification based on mathematical morphology and machine learning,'' in \emph{IEEE Power \& Energy Society Innovative Smart Grid Technologies Conference (ISGT)}, 2022, pp. 1--5.

\bibitem{lotfifard2011voltage}
S.~Lotfifard, M.~Kezunovic, and M.~J. Mousavi, ``Voltage sag data utilization for distribution fault location,'' \emph{IEEE Transactions on Power Delivery}, vol.~26, no.~2, pp. 1239--1246, 2011.

\bibitem{dong2013enhancing}
Y.~Dong, C.~Zheng, and M.~Kezunovic, ``Enhancing accuracy while reducing computation complexity for voltage-sag-based distribution fault location,'' \emph{IEEE Transactions on Power Delivery}, vol.~28, no.~2, pp. 1202--1212, 2013.

\bibitem{trindade2013fault}
F.~C. Trindade, W.~Freitas, and J.~C. Vieira, ``Fault location in distribution systems based on smart feeder meters,'' \emph{IEEE transactions on Power Delivery}, vol.~29, no.~1, pp. 251--260, 2013.

\bibitem{levie2017cayleynets}
R.~Levie, F.~Monti, X.~Bresson, and M.~M. Bronstein, ``Cayleynets: Graph convolutional neural networks with complex rational spectral filters. corr abs/1705.07664 (2017),'' \emph{arXiv preprint arXiv:1705.07664}, 2017.

\bibitem{li2015gated}
Y.~Li, D.~Tarlow, M.~Brockschmidt, and R.~Zemel, ``Gated graph sequence neural networks,'' \emph{arXiv preprint arXiv:1511.05493}, 2015.

\bibitem{velivckovic2017graph}
P.~Veli{\v{c}}kovi{\'c}, G.~Cucurull, A.~Casanova, A.~Romero, P.~Lio, and Y.~Bengio, ``Graph attention networks,'' \emph{arXiv preprint arXiv:1710.10903}, 2017.

\bibitem{kersting1991radial}
W.~H. Kersting, ``Radial distribution test feeders,'' \emph{IEEE Transactions on Power Systems}, vol.~6, no.~3, pp. 975--985, 1991.

\bibitem{dugan2011open}
R.~C. Dugan and T.~E. McDermott, ``An open source platform for collaborating on smart grid research,'' in \emph{IEEE power and energy society general meeting}, 2011, pp. 1--7.

\bibitem{kipf2016semi}
T.~N. Kipf and M.~Welling, ``Semi-supervised classification with graph convolutional networks,'' \emph{arXiv preprint arXiv:1609.02907}, 2016.

\bibitem{cai2020note}
C.~Cai and Y.~Wang, ``A note on over-smoothing for graph neural networks,'' \emph{arXiv preprint arXiv:2006.13318}, 2020.

\bibitem{ba2016layer}
J.~L. Ba, J.~R. Kiros, and G.~E. Hinton, ``Layer normalization,'' \emph{arXiv preprint arXiv:1607.06450}, 2016.

\bibitem{loshchilov2017decoupled}
I.~Loshchilov and F.~Hutter, ``Decoupled weight decay regularization,'' \emph{arXiv preprint arXiv:1711.05101}, 2017.

\bibitem{ying2019gnnexplainer}
Z.~Ying, D.~Bourgeois, J.~You, M.~Zitnik, and J.~Leskovec, ``Gnnexplainer: Generating explanations for graph neural networks,'' \emph{Advances in neural information processing systems}, vol.~32, 2019.

\bibitem{palmintier2020smart}
B.~Palmintier and B.-M. Hodge, ``{SMART-DS}: Synthetic models for advanced, realistic testing: Distribution systems and scenarios,'' National Renewable Energy Lab.(NREL), Golden, CO (United States), Tech. Rep., 2020.

\bibitem{stifter2018real}
M.~Stifter, J.~Cordova, J.~Kazmi, and R.~Arghandeh, ``Real-time simulation and hardware-in-the-loop testbed for distribution synchrophasor applications,'' \emph{Energies}, vol.~11, no.~4, p. 876, 2018.

\bibitem{liao2016micro}
A.~L. Liao, E.~M. Stewart, and E.~C. Kara, ``Micro-synchrophasor data for diagnosis of transmission and distribution level events,'' in \emph{2016 IEEE/PES Transmission and Distribution Conference and Exposition (T\&D)}.\hskip 1em plus 0.5em minus 0.4em\relax IEEE, 2016, pp. 1--5.

\bibitem{pignati2016fault}
M.~Pignati, L.~Zanni, P.~Romano, R.~Cherkaoui, and M.~Paolone, ``Fault detection and faulted line identification in active distribution networks using synchrophasors-based real-time state estimation,'' \emph{IEEE Transactions on Power Delivery}, vol.~32, no.~1, pp. 381--392, 2016.

\bibitem{malandra2018impact}
F.~Malandra, R.~Pourramezan, H.~Karimi, and B.~Sans{\`o}, ``Impact of {PMU} and smart meter applications on the performance of {LTE}-based smart city communications,'' in \emph{2018 IEEE 29th Annual International Symposium on Personal, Indoor and Mobile Radio Communications (PIMRC)}.\hskip 1em plus 0.5em minus 0.4em\relax IEEE, 2018, pp. 1--6.

\end{thebibliography}
\bibliographystyle{IEEEtran}
\vspace{-1em}
\end{document}